\documentclass[journal]{IEEEtran}
\usepackage{amsmath,amssymb,booktabs,cite,graphicx,multirow,array,tabularx,algorithm,algorithmic}
\usepackage[table]{xcolor}
\usepackage{url}
\usepackage[hidelinks]{hyperref}
\usepackage{placeins,stfloats}
\renewcommand{\thesection}{\arabic{section}}
\renewcommand{\thesubsection}{\thesection.\arabic{subsection}}
\renewcommand{\thesubsectiondis}{\thesubsection}
\renewcommand{\thetable}{\arabic{table}}
\makeatletter
\renewcommand{\@IEEEfigurecaptionsepspace}{\vskip4pt\relax}
\makeatother
\newcolumntype{Y}{>{\centering\arraybackslash}X}
\newcolumntype{P}[1]{>{\raggedright\arraybackslash}p{#1}}
\newcommand{\rev}[1]{#1}
\newcommand{\focus}[1]{#1}
\newcommand{\focusmath}{}
\newcommand{\markcolor}{}

\newcommand{\tablefont}{\fontsize{8.6}{10}\selectfont\setlength{\tabcolsep}{3pt}\renewcommand{\arraystretch}{1.09}}
\newcommand{\tablefontplain}{\tablefont\setlength{\tabcolsep}{2pt}}
\newcommand{\tablefontdense}{\tablefont\setlength{\tabcolsep}{1.5pt}}
\newcommand{\method}{FMRR}
\newcommand{\best}[1]{\textbf{#1}}
\newcommand{\second}[1]{\underline{#1}}
\newcommand{\rankbest}[1]{\best{#1}}
\newcommand{\ranksecond}[1]{\second{#1}}

\newcommand{\vresult}[1]{#1}
\definecolor{fmrrrow}{RGB}{232,240,250}

\newcommand{\VOneParamOverhead}{0.249}
\newcommand{\VOneLSParamOverhead}{0.152}
\newcommand{\VOneDFMParamOverhead}{0.132}

\newcommand{\WideTable}[1]{\begin{table*}[!tbp]\centering #1\end{table*}}

\let\TopFigure\WideFigure

\newenvironment{columntable}{\begin{table}[!htbp]\centering}{\end{table}}
\newenvironment{columnfigure}{\begin{figure}[!htbp]\centering}{\end{figure}}
\newcommand{\FirstColumnFigure}[2]{\begin{columnfigure}#1\end{columnfigure}#2}
\newcommand{\WideBlocks}[1]{#1}
\newcommand{\WideIntro}[1]{#1\par}
\newcommand{\TopColumns}[3]{#1#2#3}
\newcommand{\AlignedColumns}[2]{#1#2}

\newcommand{\CompactAppendixSections}{}
\hypersetup{hidelinks,pdftitle={Interference-Aware Frequency-Domain Fusion for Multi-Modal Saliency Detection},pdfauthor={Haifa Zhang, Yijing Wang, Zhiqiang Zuo},pdfsubject={Research preprint}}
\title{Interference-Aware Frequency-Domain Fusion for Multi-Modal Saliency Detection}
\author{Haifa~Zhang, Yijing~Wang, and Zhiqiang~Zuo%
\thanks{Tianjin Key Laboratory of Intelligent Unmanned Swarm Technology and System, School of Electrical and Information Engineering, Tianjin University, Tianjin 300072, China.}}
\begin{document}
\maketitle
\thispagestyle{plain}
\pagestyle{plain}
\begin{abstract}
\markcolor RGB-depth (RGB-D) and RGB-thermal (RGB-T) salient object detection (SOD) exploit complementary sensing, but degraded inputs can introduce unreliable structures and misleading responses into cross-modal fusion. We address this interference with Frequency-Domain Modality Reliability Routing (FMRR), \focus{a lightweight, architecture-preserving pre-fusion contribution-routing interface for existing bimodal SOD detectors.} \focus{FMRR jointly observes normalized Fourier-amplitude statistics from both inputs and allocates two relative participation weights under a fixed budget. The weights scale the complete feature hierarchies, while sample-level modality dropout (MD) trains one shared detector for complete, RGB-only, and auxiliary-only inputs via the original SOD objective.} \focus{Matched comparisons show that combining MD and FMRR improves complete-input prediction and single-modality fallback. Under identical MD training, FMRR comes within 0.010 $S_m$ of an availability-flag hard gate on missing inputs and improves over alternative mechanisms on nonzero and degraded inputs. Its weights track degradation severity and the per-sample advantage of independently trained single-modality detectors.} Experiments cover three RGB-T and five RGB-D benchmarks using ConTriNet, LSNet, DFM-Net, and VST-T++. \focus{On ConTriNet, the complete-input RGB-D means are 0.901 $S_m$, 0.942 adaptive E-measure, 0.898 adaptive F-measure, and 0.037 MAE, with at most 249,458 additional parameters.} These results show that regulating modality contributions before fusion improves prediction while suppressing interference from unreliable inputs.
\end{abstract}
\begin{IEEEkeywords}
Multi-modal salient object detection; RGB-D salient object detection; RGB-T salient object detection; \rev{modality reliability routing}; \rev{frequency-domain fusion}; \rev{interference suppression}
\end{IEEEkeywords}
\section{Introduction}
\label{loc:intro}
\rev{Salient object detection (SOD) predicts the image regions that attract human attention and supports segmentation, tracking, image editing, and robotic perception. RGB-depth (RGB-D) and RGB-thermal (RGB-T) SOD complement visible appearance with geometric or thermal evidence, improving saliency prediction in scenes where one sensor alone is ambiguous~\cite{zhou2021rgbdsurvey,tu2023rgbtbenchmark,pang2023caver}. This benefit depends on the evidence carried by each stream: depth may reveal object boundaries that are weak in RGB, while thermal sensing may retain foreground contrast under poor illumination.}

\rev{Most bimodal detectors strengthen the way two representations interact through attention, multi-scale aggregation, and specialized decoding~\cite{tu2021midd,zhou2022ecffnet,wang2022cgfnet,liu2022swinnet,pang2023caver}. Yet the value of a stream varies from sample to sample. A depth map with missing measurements, a thermal image dominated by background heat, or a weak RGB observation can inject responses that are repeatedly propagated by the subsequent fusion hierarchy. This motivates assessing the contribution of each modality before fusion, so that unreliable evidence is attenuated before it propagates through the detector. Quality-aware feature manipulation and learned multimodal gating provide related ways to regulate modality contributions~\cite{zhang2021dqfm,arevalo2017gmu}.}

\rev{We propose FMRR as a \focus{lightweight, architecture-preserving pre-fusion contribution-routing interface} rather than redesigning existing detectors or replacing native fusion. Its attachment point follows the host architecture's feature-flow topology. Joint modality observations produce fixed-budget, task-guided participation weights before detector-specific interaction. Fig.~\ref{fig:overview_a0_a3} compares the baseline and routed detector under complete and single-modality inputs.}

\rev{FMRR is trained with sample-level modality dropout (MD), which exposes the shared detector to complete, RGB-only, and auxiliary-only states. Modality dropout has been used to improve prediction with missing channels~\cite{neverova2016moddrop}; here, it is coupled with an explicit routing pathway before fusion. The original SOD objective trains the routing weights through their effect on the final prediction, without sensor-quality annotations or an auxiliary routing loss. This coupling connects the availability changes seen during training to the suppression of unreliable feature responses.}

\rev{FMRR instantiates this interface by normalizing each input, summarizing its filtered Fourier-amplitude distribution, jointly observing the descriptors, and allocating a fixed participation budget. A coupled softmax makes the shares relative. They scale the hierarchies before native interaction, while the original SOD objective supplies feedback through the final prediction. Identity-initialized routing preserves the native encoders, fusion, decoder, and objective. Experiments test the demand--pathway mechanism, graded and real modality changes, and transfer across native fusion designs.}

\FirstColumnFigure{
\centering
\includegraphics[width=\linewidth]{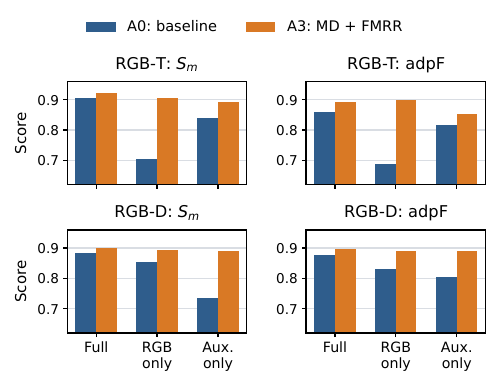}
\caption{\rev{Matched A0 and A3 behavior for ConTriNet under complete, RGB-only, and auxiliary-only inference. Bars are five-seed test-set means over the corresponding RGB-T or RGB-D benchmarks.}}
\label{fig:overview_a0_a3}
}{

The main contributions are summarized as follows:
\begin{itemize}
    \item \rev{We formulate modality robustness as \focus{task-guided pre-fusion contribution routing}, learning sample-dependent relative modality participation from joint observations, controlled availability changes, and the native SOD objective.}
    \item \rev{We develop FMRR as a \focus{lightweight, architecture-preserving plug-in interface} that applies fixed-budget relative weights to complete modality feature hierarchies before each detector's native interaction, while retaining its original encoders, fusion modules, decoder, and task objective.}
    \item \rev{We validate the formulation through matched factorial and admission-mechanism comparisons, graded and real degradation analyses, external single-modality references, and integrations with two RGB-T detectors and dedicated RGB-D detectors.}
\end{itemize}
}

\section{Related Work}
\label{loc:related}
\subsection{RGB-T Salient Object Detection}
\label{loc:rgbt_related}
RGB-T SOD combines visible and thermal infrared images. Early RGB-T benchmarks include VT821~\cite{wang2018rgbt} and VT1000~\cite{tu2020rgbt}, while VT5000~\cite{tu2023rgbtbenchmark} provides a larger-scale benchmark with more diverse scenes. Traditional approaches often relied on graph models or hand-crafted cross-modal cues. Modern deep networks learn modality-specific and cross-modal features in an end-to-end manner~\cite{tu2021midd,zhou2022ecffnet,wang2022cgfnet,pang2023caver,tang2025contrinet,zhou2023lsnet}.

The main design question in RGB-T SOD is how to exploit the complementary properties of the two modalities. RGB images preserve texture and semantic appearance, but fail in low-light or visually confusing environments. Thermal images are more robust to illumination, but usually lack fine-grained texture and may introduce ambiguous heat responses. Recent methods therefore design fusion modules that balance these streams, including CNN-based dual-stream interaction, one-stream refinement, transformer-based cross-modal mixing, edge-aware designs, and progressive cross-modal cross-scale fusion~\cite{liu2022swinnet,pang2023caver,tang2025contrinet}. ConTriNet~\cite{tang2025contrinet} follows a divide-and-conquer principle. It uses a unified encoder and three flows to separately explore RGB-specific, thermal-specific, and complementary information. Its Modality-induced Feature Modulator, Residual Atrous Spatial Pyramid Module, and Modality-aware Dynamic Aggregation Module are designed to reduce modality discrepancy and aggregate saliency cues. LSNet~\cite{zhou2023lsnet}, in contrast, focuses on efficiency. It uses a lightweight backbone and transfer-based training mechanisms to improve feature quality without increasing inference cost.

\rev{These methods make RGB-T interaction increasingly expressive. FMRR complements them with an earlier participation decision: the observation branch reads the raw inputs, while the allocated shares regulate the resulting feature hierarchies before native fusion. The encoder inputs and the backbone-specific interaction pathway remain unchanged.}

\subsection{RGB-D Salient Object Detection}
\label{loc:rgbd_related}
RGB-D SOD has been studied with datasets such as NJU2K~\cite{ju2014nju2k}, NLPR~\cite{peng2014nlpr}, DES~\cite{cheng2014des}, LFSD~\cite{li2014lfsd}, and SIP~\cite{fan2021sip}. Depth cues can help separate foreground and background, especially when RGB texture is confusing. However, depth quality varies across sensors and scenes. A shallow object, reflective surface, or missing depth region can provide misleading geometric evidence. RGB-D SOD therefore needs not only mechanisms that exploit depth when it is useful, but also mechanisms that reduce its influence when it is unreliable. Representative RGB-D SOD models have explored depth quality assessment, depth distillation, cross-modal weighting, selective modulation, hierarchical interaction, and depth-aware fusion~\cite{zhang2021dqfm,fu2020jldcf,liu2020ssma,wu2023hidanet}.

\rev{The RGB-D and RGB-T settings share the same high-level requirement: complementary evidence should be admitted in a sample-dependent manner. RGB-D methods such as JL-DCF, SSMA, DFM-Net, SPNet, CIR-Net, CAVER, and VST++ develop increasingly capable interaction and quality-aware feature processing~\cite{fu2020jldcf,liu2020ssma,zhang2021dqfm,zhou2021spnet,cong2022cirnet,pang2023caver,liu2024vstpp}. We retain these native interactions and place a shared participation interface immediately before them.}

\subsection{Robust and Modality-Incomplete Learning}
\label{sec:robust_related}
\rev{Robust multimodal learning studies complete, degraded, and missing inputs. RGB-D quality-aware models selectively manipulate depth features~\cite{zhang2021dqfm,wu2022robustrgbd}; GRNet develops a generalizable cross-modal interaction design for anti-interference learning~\cite{wen2023grnet}; UniMRSeg reconstructs and aligns incomplete representations~\cite{zhao2025unimrseg}; and CoLA combines a language-based quality assessor with Conditional Dropout~\cite{hao2024cola}. \focus{Our formulation adopts sample-level availability intervention as the demand, a modality-level path before native fusion, and the original SOD objective as feedback.} The matched comparisons in Sec.~\ref{sec:admission_comparison} hold the intervention and backbone fixed when evaluating alternative admission mechanisms.}

\subsection{Frequency-Domain Observation for Modality Routing}
\label{loc:frequency_related}
\rev{Frequency-domain representations provide a compact image-wide description and have been used for feature learning and cross-modal fusion~\cite{jin2022fcmnet,lyu2026freqsal}. \focus{FMRR uses normalized log-amplitude statistics only to allocate image-wide modality participation. Because Fourier amplitude is invariant to an ideal circular translation, FMRR does not estimate pure relative displacement; the detector's native fusion pathway remains responsible for spatial correspondence.} Sec.~\ref{sec:estimator_ablation} compares this observation space with spatial statistics, a hand-crafted quality vector, and a parameter-matched CNN under the same participation interface. Appendix~E evaluates this division of roles under relative auxiliary-stream shifts.}

\section{Proposed Method}
\subsection{Problem Formulation and Overview}
\label{loc:formulation}
\rev{Let $x^{r}\in\mathbb{R}^{C_r\times H\times W}$ denote an RGB input and $x^{a}\in\mathbb{R}^{C_a\times H\times W}$ denote the auxiliary depth or thermal input. A baseline detector extracts feature hierarchies $\{f_l^r\}_{l=1}^{L}$ and $\{f_l^a\}_{l=1}^{L}$, combines them through a native cross-modal pathway $\Phi$, and predicts a saliency map with decoder $D$:}
{\markcolor
\begin{equation}
f_l^m=E_m^l(x^m),\qquad
\hat{y}=D\!\left(\Phi(\{f_l^r\}_{l=1}^{L},\{f_l^a\}_{l=1}^{L})\right),
\end{equation}
}

\rev{\focus{We introduce a participation interface before $\Phi$. FMRR jointly predicts two non-negative shares from the raw inputs and preserves a fixed total budget:}}
{\markcolor
\begin{gather}
\focusmath
 w^r+w^a=2,\qquad
 \bar f_l^m=w^m f_l^m,
 \label{eq:fixed_budget}\\
\hat y=D\!\left(\Phi(\{\bar f_l^r\}_{l=1}^{L},\{\bar f_l^a\}_{l=1}^{L})\right).
\end{gather}
}
\rev{\focus{The identity allocation is $(w^r,w^a)=(1,1)$. Increasing one share reduces the other under the fixed budget. Here, task utility refers to a modality's sample-dependent predictive usefulness for SOD. The routing weights express task-guided relative participation and are learned through the native SOD objective (Fig.~\ref{fig:fmrr}).}}

\subsection{Modality Dropout}
\label{loc:md}
\rev{Sample-level modality dropout creates the controlled intervention used to guide participation. For each sample $i$, $d_i\sim\mathrm{Bernoulli}(p)$ decides whether a stream is removed, and $q_i\sim\mathrm{Bernoulli}(0.5)$ selects the stream to remove:}
{\markcolor
\begin{equation}
(\tilde{x}^{r}_{i}, \tilde{x}^{a}_{i}) =
\begin{cases}
(0, x^{a}_{i}), & d_i=1, q_i=0,\\
(x^{r}_{i}, 0), & d_i=1, q_i=1,\\
(x^{r}_{i}, x^{a}_{i}), & d_i=0.
\end{cases}
\end{equation}
}
\rev{The main configuration uses $p=0.5$, which preserves complete samples while regularly exposing the network to two controlled contrasting states. FMRR turns this intervention into a learnable participation signal: it attenuates the feature hierarchy generated by the unavailable stream and preserves the contribution of the available stream before repeated cross-modal interaction. This explicit control adjusts feature contributions with input availability while preserving the native fusion operations.}

\rev{This intervention is deliberately sensor-agnostic: it supplies a clear contrast between useful and unavailable streams without introducing corruption labels or a separate quality objective. The learned policy is subsequently evaluated on nonzero substitutes, graded local degradations, and real sensor states.}

\subsection{\rev{Frequency-Domain Modality Reliability Routing (FMRR)}}
\label{loc:router}
\rev{\focus{For each modality $m\in\{r,a\}$, FMRR first applies per-image, per-channel normalization:}}
{\markcolor
\begin{equation}
\focusmath
z_c^m=\frac{\tilde x_c^m-\mu_{hw}(\tilde x_c^m)}
{\sqrt{\sigma_{hw}^2(\tilde x_c^m)+\epsilon}}.
\label{eq:input_normalization}
\end{equation}
}
\rev{\focus{Here $\mu_{hw}$ and $\sigma_{hw}^2$ are the spatial mean and variance of each channel, and $\epsilon>0$ stabilizes the denominator. Mean subtraction removes global offsets, while variance normalization reduces sensitivity to input scale. Appendix~E evaluates photometric stability and relative spatial shifts. A real two-dimensional Fourier transform then produces the log-amplitude representation:}}
{\markcolor
\begin{gather}
A^m=\log\!\left(1+\left|\mathcal F_{rfft2}(z^m)\right|\right).\\
\hat A^m=A^m\odot G^m,
\end{gather}
}
\rev{where $G^m$ is a learnable channel-wise spectral filter. It is bilinearly resized when the test resolution differs from the initialized spectral size. This filter operates in the observation branch; the native encoders receive the detector inputs directly. Channel-wise mean, standard deviation, and maximum over the frequency grid form the descriptor:}
{\markcolor
\begin{equation}
s^m=\left[\mu(\hat A^m),\,\sigma(\hat A^m),\,\max(\hat A^m)\right]\in\mathbb R^{3C_m}.
\end{equation}
}
\rev{\focus{The two descriptors are concatenated so that participation is allocated from their joint state rather than predicted independently. The concatenated descriptor has $3(C_r+C_a)$ entries. A shared MLP maps it to two logits, with one output for each modality:}}
{\markcolor
\begin{gather}
\focusmath
q=[s^r;s^a],\qquad
u=W_2\,\delta(W_1q+b_1)+b_2,\\
\focusmath
\begin{bmatrix}w^r\\w^a\end{bmatrix}
=2\,\mathrm{softmax}(u).
\label{eq:joint_allocator}
\end{gather}
}
\rev{\focus{Here $W_1\in\mathbb R^{h\times 3(C_r+C_a)}$ and $W_2\in\mathbb R^{2\times h}$ are the MLP weight matrices; $b_1\in\mathbb R^h$ and $b_2\in\mathbb R^2$ are their biases. The hidden width is $h$, and $\delta$ is ReLU. The final layer has zero-initialized weights and bias, so Eq.~\eqref{eq:joint_allocator} begins exactly at $(1,1)$. The fixed budget gives each coefficient a direct relative interpretation: $w^r>1$ reallocates participation toward RGB, whereas $w^a>1$ reallocates it toward the auxiliary stream. The same share is applied to every level of a modality hierarchy, creating one coherent admission decision before repeated native fusion.}}

\TopFigure{
\centering
% User-revised editable PowerPoint framework, exported as a cropped vector PDF.
\includegraphics[width=\textwidth]{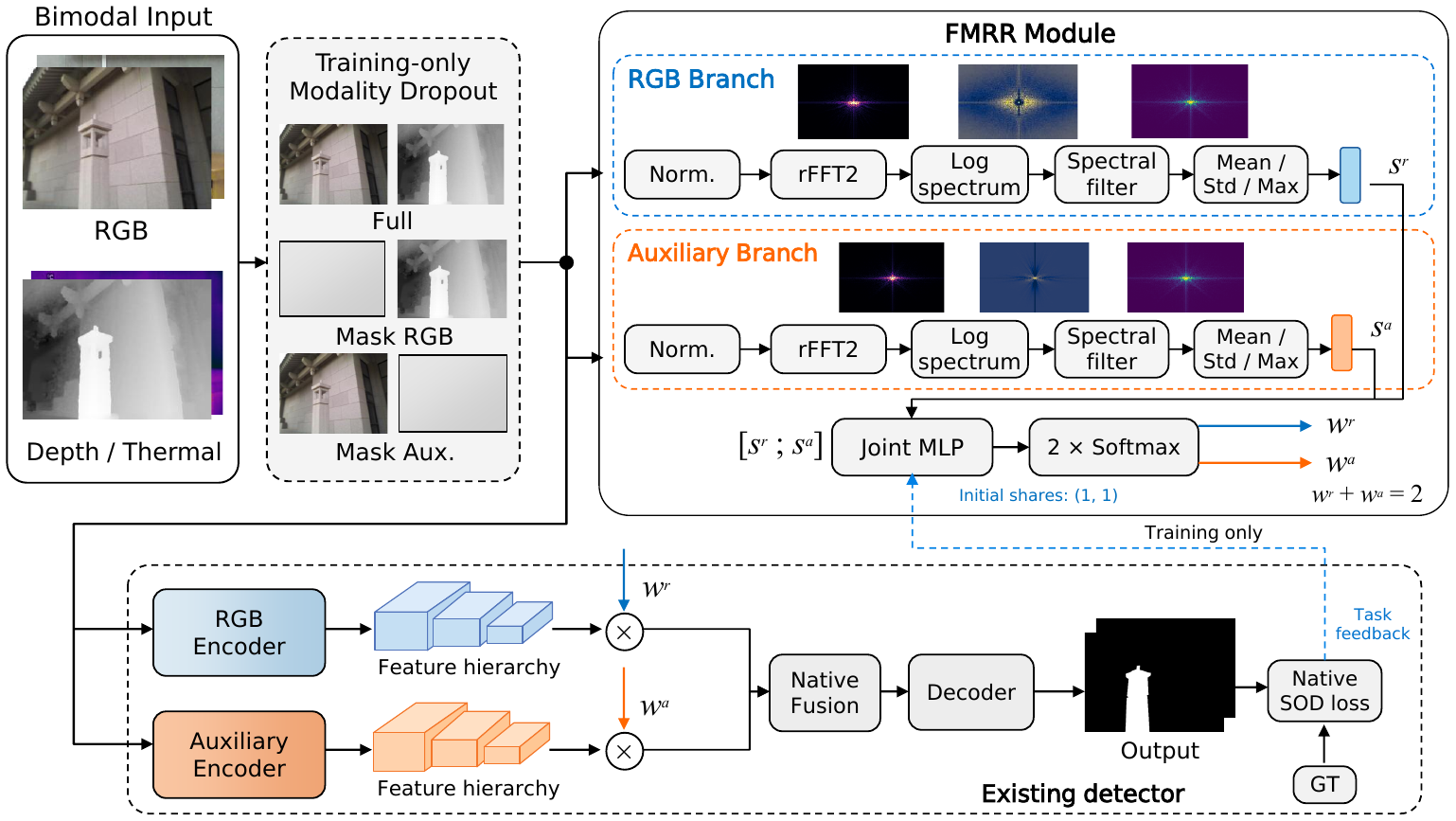}
\caption{\rev{Overview of guided pre-fusion modality participation. MD-selected inputs feed the spectral observation branches and native encoders in parallel. \focus{FMRR jointly allocates two shares with $w^r+w^a=2$ and scales each modality's complete feature hierarchy before native fusion.} The dashed feedback path represents supervision from the native SOD objective through the detector.}}
\label{fig:fmrr}
\vspace{6pt}
}

\subsection{\rev{Integration Into Bimodal Detectors}}
\label{loc:integration}
\rev{For ConTriNet, the shares scale every element of the Res2Net-based RGB and auxiliary feature tuples before the original Modality-induced Feature Modulator and subsequent aggregation modules~\cite{gao2021res2net}. For LSNet, they scale all five MobileNetV2 feature levels before native fusion~\cite{sandler2018mobilenetv2}. These two integrations evaluate the same pathway in strong and lightweight RGB-T/RGB-D architectures.}

\rev{\focus{We also integrate FMRR into the dedicated RGB-D detectors DFM-Net~\cite{zhang2021dqfm} and VST-T++~\cite{liu2024vstpp}. The shares scale five RGB/depth hierarchies before DFM-Net's depth-quality processing and additive fusion, and the token hierarchies before VST-T++ cross-modal interaction. These integrations retain the native encoders, interaction modules, decoders, and objectives while validating topology-compatible placement: the routing principle is shared, but its attachment point and routed hierarchy follow each architecture's native feature flow (Appendix~D), without redesigning fusion.}}

\subsection{Training Objective and Participation Learning}
\label{loc:objective}
\rev{\focus{FMRR learns participation from the native task supervision without explicit sensor-quality annotations, corruption labels, or an auxiliary routing loss. Let $\mathcal{L}_{base}$ denote the complete objective of the selected detector, including its main and original side-output terms. Training retains:}}
{\markcolor
\begin{equation}
\mathcal{L}_{total} = \mathcal{L}_{base}.
\end{equation}
}
\rev{\focus{Because $w^m$ scales every level, the chain rule aggregates feedback from this unchanged task objective across the complete hierarchy:}}
{\markcolor
\begin{equation}
\focusmath
\frac{\partial \mathcal{L}_{total}}{\partial w^m}
=\sum_{l=1}^{L}\left\langle
\frac{\partial \mathcal{L}_{total}}{\partial \bar{f}_l^m},
f_l^m\right\rangle .
\label{eq:participation_gradient}
\end{equation}}
\rev{\focus{Through the coupled softmax in Eq.~\eqref{eq:joint_allocator}, the difference between the two hierarchy-level signals reallocates the fixed budget. MD makes this contrast repeatedly observable, while the original SOD loss evaluates the resulting prediction. The allocator thus adjusts relative participation through prediction-loss feedback.}}

\subsection{Plug-in Complexity}
\label{loc:complexity}
\rev{FMRR adds two spectral filters and one joint two-layer allocator. For modality channel counts $C_r$ and $C_a$, spectral size $H\times(\lfloor W/2\rfloor+1)$, and hidden dimension $h$, its parameter count is:}
{\markcolor
\begin{equation}
\begin{aligned}
N={}&(C_r+C_a)H(\lfloor W/2\rfloor+1)\\
   &+3h(C_r+C_a)+3h+2.
\end{aligned}
\end{equation}
}
\rev{The first term contains the learnable filters and the remaining terms contain the joint MLP. FMRR adds 249,458 parameters (\VOneParamOverhead\,M) for the $352\!\times\!352$, 3+1-channel ConTriNet configuration, 152,210 parameters (\VOneLSParamOverhead\,M) for the $224\!\times\!224$, 3+3-channel LSNet/VST-T++ configurations, and 132,338 parameters (\VOneDFMParamOverhead\,M) for the $256\!\times\!256$, 3+1-channel DFM-Net configuration. Table~\ref{tab:parameter_accounting} reports these increments together with the original and resulting total detector sizes. The encoders, interaction modules, and decoder remain unchanged.}

% Original detector parameters, FMRR increments, and resulting totals.
\begin{columntable}
\caption{Parameter accounting for the four integration backbones. Original denotes the complete baseline detector, added denotes FMRR alone, and total is their sum.}
\label{tab:parameter_accounting}
\centering
\tablefontdense
\begin{tabular*}{\linewidth}{@{\extracolsep{\fill}}lrrrr@{}}
\toprule
Detector & \shortstack{Original\\(M)} & \shortstack{Added\\(M)} & \shortstack{Total\\(M)} & \shortstack{Increase\\(\%)} \\
\midrule
ConTriNet~\cite{tang2025contrinet} & 35.070 & 0.249 & 35.319 & 0.711 \\
LSNet~\cite{zhou2023lsnet} & 5.390 & 0.152 & 5.542 & 2.824 \\
DFM-Net~\cite{zhang2021dqfm} & 2.217 & 0.132 & 2.350 & 5.968 \\
VST-T++~\cite{liu2024vstpp} & 100.510 & 0.152 & 100.662 & 0.151 \\
\bottomrule
\end{tabular*}
\end{columntable}

\begin{algorithm}[tbp]
\markcolor
\caption{\rev{Joint FMRR Training and Inference}}
\label{alg:fmrr}
\begin{algorithmic}[1]
\REQUIRE RGB input $x^r$, auxiliary input $x^a$, detector encoders $E_r,E_a$, fusion $\Phi$, decoder $D$
\IF{training}
    \STATE Apply sample-level modality dropout to obtain $\tilde{x}^r,\tilde{x}^a$
\ELSE
    \STATE Set $\tilde{x}^r=x^r$ and $\tilde{x}^a=x^a$ unless a robustness mode explicitly drops one modality
\ENDIF
\FOR{$m\in\{r,a\}$}
    \STATE \focus{Normalize $\tilde{x}^m$ per image and channel to obtain $z^m$}
    \STATE Compute $A^m=\log(1+|\mathcal{F}_{rfft2}(z^m)|)$ and filter it with $G^m$
    \STATE Pool mean, standard deviation, and maximum statistics into $s^m$
\ENDFOR
\STATE \focus{Jointly allocate $[w^r,w^a]=2\,\mathrm{softmax}(\mathrm{MLP}([s^r;s^a]))$}
\STATE Extract feature pyramids $\{f_l^r\}=E_r(\tilde{x}^r)$ and $\{f_l^a\}=E_a(\tilde{x}^a)$
\STATE Reweight features $\bar{f}_l^r=w^rf_l^r$ and $\bar{f}_l^a=w^af_l^a$
\STATE Predict saliency map $\hat{y}=D(\Phi(\{\bar{f}_l^r\},\{\bar{f}_l^a\}))$
\end{algorithmic}
\end{algorithm}

\section{Experiments}
\subsection{Datasets and Protocol}
\label{loc:protocol}
\rev{\focus{RGB-T experiments follow the ConTriNet protocol: 2,500 image pairs of VT5000~\cite{tu2023rgbtbenchmark} are used for training, and testing uses the remaining 2,500 VT5000 pairs together with VT821~\cite{wang2018rgbt} and VT1000~\cite{tu2020rgbt}. RGB-D experiments train on the training portions of NJU2K~\cite{ju2014nju2k} and NLPR~\cite{peng2014nlpr} and test on NJU2K, NLPR, DES~\cite{cheng2014des}, LFSD~\cite{li2014lfsd}, and SIP~\cite{fan2021sip}; DFM-Net and VST-T++ follow the same RGB-D split. Neither protocol defines a separate validation set, so all benchmark and mechanism analyses are reported on these test sets. Within each backbone, settings share the data split, training duration, final-epoch checkpoint rule, and evaluation procedure. One checkpoint is evaluated under complete input, RGB-only input, auxiliary-only input, nonzero substitutes, local masks, and held-out sensor degradations. Local holes and masks are sampled independently of saliency ground truth.}}

\subsection{Evaluation Metrics}
\label{loc:metrics}
\rev{We report MAE, structure measure $S_m$, adaptive E-measure (adpE), adaptive F-measure (adpF), and weighted F-measure $F_\beta^w$~\cite{margolin2014evaluate,fan2021smeasure,fan2018emeasure}. Lower MAE is better; higher values are better for the remaining metrics. $S_m$ is the primary metric in matched analyses, while $F_\beta^w$ and the foreground statistics in Appendix~C characterize single-stream structural recovery.}

\subsection{Implementation Details}
\label{loc:implementation}
% Author-supplied protocol statement; training provenance is tracked separately from layout checks.
\rev{\focus{Every detector in the main benchmark Tables~\ref{tab:rgbt_sota_full}--\ref{tab:vstpp_transfer} was reproduced under its documented architecture-specific protocol; no published final checkpoint is used. ConTriNet, LSNet, DFM-Net, and VST-T++ also support the controlled factorial, routing, and transfer analyses.}}

\rev{Specifically, ConTriNet uses $352\times352$ inputs and a 100-epoch training schedule, while LSNet uses $224\times224$ inputs and a 20-epoch schedule. DFM-Net and VST-T++ retain their respective $256\times256$ and $224\times224$ RGB-D configurations. Unless otherwise specified, MD-enabled settings use a dropout probability of $p=0.5$. \focus{For A1, some runs at the original rates showed non-finite gradients or nonconvergence; the five reported converged runs use reduced, backbone-specific learning rates. All five A3 runs converged at the original rates.} The FMRR allocator uses a 16-dimensional hidden layer with zero-initialized output weights and biases, giving equal initial participation shares of $(1,1)$. For each run, the final-epoch checkpoint is evaluated under complete, single-modality, and degraded-input conditions without additional fine-tuning. The native task objective of each detector is preserved throughout training.}

\subsection{\rev{Complete Input and Cross-Architecture Transfer}}
\label{sec:complete_results}
\rev{The complete-input comparison tests whether interference-aware routing improves different native fusion designs. Tables~\ref{tab:rgbt_sota_full} and~\ref{tab:rgbd_sota_full} anchor this comparison on the RGB-T and RGB-D test protocols. On RGB-T, the coupled model improves ConTriNet on every dataset and metric; on RGB-D, it improves the five-dataset mean of every metric and lowers MAE on every dataset. LSNet improves on all three RGB-T datasets and obtains larger RGB-D gains. Table~\ref{tab:vstpp_transfer} extends the identical interface to lightweight DFM-Net and transformer-based VST-T++; against baselines trained under the same protocol, the routed models improve $S_m$ and MAE on all five RGB-D benchmarks. The added parameters account for 5.968\% and 0.151\% of their respective original detector sizes, as detailed in Table~\ref{tab:parameter_accounting}. Appendix~D further confirms the same complete/nonzero/graded advantage under the full matched state protocol.}

\WideTable{
\caption{Full-modality RGB-T SOD comparison. \focus{Learning-based detectors follow Sec.~4.1 (five-seed means); MTMR, M3S-NIR, and SGDL use their released settings.} Bold/underline denotes best/second-best; ties at the reported precision share the same mark. ``+FMRR (Ours)'' is the complete MD--FMRR configuration.}
\label{tab:rgbt_sota_full}
\centering
\tablefontplain
\begin{tabular*}{\textwidth}{@{\extracolsep{\fill}}lcccccccccccc@{}}
\toprule
Method & \multicolumn{4}{c}{VT821} & \multicolumn{4}{c}{VT1000} & \multicolumn{4}{c}{VT5000} \\
\cmidrule(lr){2-5}\cmidrule(lr){6-9}\cmidrule(lr){10-13}
 & $S_m\uparrow$ & adpE$\uparrow$ & adpF$\uparrow$ & MAE$\downarrow$ & $S_m\uparrow$ & adpE$\uparrow$ & adpF$\uparrow$ & MAE$\downarrow$ & $S_m\uparrow$ & adpE$\uparrow$ & adpF$\uparrow$ & MAE$\downarrow$ \\
\midrule
S2MA~\cite{liu2020ssma} & 0.826 & 0.826 & 0.714 & 0.083 & 0.910 & 0.911 & 0.841 & 0.031 & 0.846 & 0.864 & 0.744 & 0.058 \\
JL-DCF~\cite{fu2020jldcf} & 0.835 & 0.823 & 0.719 & 0.079 & 0.903 & 0.893 & 0.826 & 0.033 & 0.854 & 0.853 & 0.733 & 0.052 \\
HAIN~\cite{li2021hainet} & 0.844 & 0.875 & 0.777 & 0.050 & 0.911 & 0.923 & 0.875 & 0.028 & 0.860 & 0.893 & 0.802 & 0.043 \\
SPNet~\cite{zhou2021spnet} & 0.854 & 0.884 & 0.808 & 0.040 & 0.910 & 0.929 & 0.872 & 0.026 & 0.861 & 0.896 & 0.829 & 0.038 \\
DCMF~\cite{wang2022dcmf} & 0.844 & 0.859 & 0.822 & 0.059 & 0.909 & 0.905 & 0.831 & 0.029 & 0.853 & 0.866 & 0.745 & 0.054 \\
CIR-Net~\cite{cong2022cirnet} & 0.851 & 0.880 & 0.773 & 0.049 & 0.906 & 0.914 & 0.854 & 0.031 & 0.865 & 0.896 & 0.788 & 0.043 \\
TBINet~\cite{wang2022tbinet} & 0.881 & 0.916 & 0.836 & 0.034 & 0.921 & 0.934 & 0.894 & 0.024 & 0.882 & 0.921 & 0.836 & 0.035 \\
RAFNet~\cite{wu2022robustrgbd} & 0.876 & 0.902 & 0.819 & 0.042 & 0.924 & 0.930 & 0.883 & 0.022 & 0.881 & 0.917 & 0.830 & 0.037 \\
PopNet~\cite{wu2023popnet} & 0.850 & 0.878 & 0.787 & 0.047 & 0.920 & 0.929 & 0.887 & 0.022 & 0.876 & 0.912 & 0.828 & 0.035 \\
HiDANet~\cite{wu2023hidanet} & 0.883 & 0.916 & 0.838 & 0.035 & 0.916 & 0.934 & 0.889 & 0.025 & 0.878 & 0.916 & 0.840 & \ranksecond{0.034} \\
MTMR~\cite{wang2018rgbt} & 0.715 & 0.809 & 0.653 & 0.112 & 0.698 & 0.825 & 0.705 & 0.121 & 0.673 & 0.790 & 0.589 & 0.115 \\
M3S-NIR~\cite{tu2019m3snir} & 0.719 & 0.851 & 0.722 & 0.141 & 0.717 & 0.819 & 0.705 & 0.147 & 0.640 & 0.775 & 0.565 & 0.172 \\
SGDL~\cite{tu2020rgbt} & 0.759 & 0.840 & 0.723 & 0.086 & 0.778 & 0.846 & 0.761 & 0.091 & 0.738 & 0.815 & 0.667 & 0.090 \\
ADF~\cite{tu2023rgbtbenchmark} & 0.802 & 0.837 & 0.710 & 0.079 & 0.905 & 0.915 & 0.842 & 0.036 & 0.858 & 0.882 & 0.767 & 0.050 \\
MIDD~\cite{tu2021midd} & 0.863 & 0.884 & 0.793 & 0.049 & 0.896 & 0.916 & 0.863 & 0.033 & 0.849 & 0.886 & 0.784 & 0.049 \\
CSRNet~\cite{huo2022ecsrnet} & 0.880 & 0.898 & 0.821 & 0.043 & 0.906 & 0.916 & 0.867 & 0.026 & 0.859 & 0.900 & 0.806 & 0.046 \\
CGFNet~\cite{wang2022cgfnet} & 0.874 & 0.907 & 0.833 & 0.038 & 0.917 & 0.931 & 0.897 & 0.028 & 0.879 & 0.910 & \ranksecond{0.848} & 0.038 \\
MMNet~\cite{gao2022uifnet} & 0.867 & 0.886 & 0.788 & 0.042 & 0.908 & 0.917 & 0.853 & 0.029 & 0.851 & 0.883 & 0.769 & 0.046 \\
MIA~\cite{liang2022mia} & 0.839 & 0.842 & 0.737 & 0.074 & 0.919 & 0.920 & 0.857 & 0.028 & 0.873 & 0.883 & 0.781 & 0.044 \\
ECFFNet~\cite{zhou2022ecffnet} & 0.868 & 0.892 & 0.800 & 0.038 & 0.912 & 0.920 & 0.867 & 0.025 & 0.870 & 0.902 & 0.802 & 0.040 \\
OSRNet~\cite{huo2022osrnet} & 0.868 & 0.885 & 0.804 & 0.046 & 0.922 & 0.932 & 0.886 & 0.025 & 0.869 & 0.903 & 0.819 & 0.042 \\
CAVER~\cite{pang2023caver} & 0.885 & 0.911 & 0.832 & 0.035 & 0.932 & 0.934 & 0.893 & 0.019 & 0.882 & 0.919 & 0.838 & 0.035 \\
\midrule
LSNet~\cite{zhou2023lsnet} & 0.877 & 0.911 & 0.827 & 0.033 & 0.924 & 0.936 & 0.887 & 0.022 & 0.876 & 0.916 & 0.827 & 0.036 \\
LSNet+\method{} (Ours) & 0.879 & 0.913 & 0.829 & \ranksecond{0.032} & 0.926 & 0.938 & 0.889 & 0.021 & 0.878 & 0.918 & 0.829 & 0.035 \\
ConTriNet~\cite{tang2025contrinet} & \ranksecond{0.891} & \ranksecond{0.918} & \ranksecond{0.839} & 0.034 & \ranksecond{0.936} & \ranksecond{0.940} & \ranksecond{0.898} & \ranksecond{0.018} & \ranksecond{0.888} & \ranksecond{0.925} & 0.844 & \ranksecond{0.034} \\
ConTriNet+\method{} (Ours) & \rankbest{0.909} & \rankbest{0.928} & \rankbest{0.864} & \rankbest{0.027} & \rankbest{0.952} & \rankbest{0.963} & \rankbest{0.927} & \rankbest{0.017} & \rankbest{0.910} & \rankbest{0.938} & \rankbest{0.888} & \rankbest{0.027} \\
\bottomrule
\end{tabular*}
}

\WideTable{
\caption{Full-modality RGB-D SOD comparison on NJU2K, DES, LFSD, NLPR, and SIP. \focus{All detectors are trained and evaluated under the protocol of Sec.~4.1; values are five-seed means. Mean is computed from unrounded per-dataset values.} Bold and underlined values indicate the best and second-best results in each metric column, respectively; ties at the reported precision share the same mark. Entries labeled ``+FMRR (Ours)'' use the complete MD--FMRR configuration.}
\label{tab:rgbd_sota_full}
\centering
\tablefontplain
\begin{tabular*}{\textwidth}{@{\extracolsep{\fill}}lcccccccccccc@{}}
\toprule
Method & \multicolumn{4}{c}{NJU2K} & \multicolumn{4}{c}{DES} & \multicolumn{4}{c}{LFSD} \\
\cmidrule(lr){2-5}\cmidrule(lr){6-9}\cmidrule(lr){10-13}
 & $S_m\uparrow$ & adpE$\uparrow$ & adpF$\uparrow$ & MAE$\downarrow$ & $S_m\uparrow$ & adpE$\uparrow$ & adpF$\uparrow$ & MAE$\downarrow$ & $S_m\uparrow$ & adpE$\uparrow$ & adpF$\uparrow$ & MAE$\downarrow$ \\
\midrule
CGFNet~\cite{wang2022cgfnet} & 0.869 & 0.888 & 0.826 & 0.080 & 0.876 & 0.956 & 0.886 & 0.061 & 0.812 & 0.847 & 0.817 & 0.128 \\
TBINet~\cite{wang2022tbinet} & 0.879 & 0.896 & 0.834 & 0.077 & 0.886 & 0.964 & 0.894 & 0.059 & 0.822 & 0.855 & 0.825 & 0.123\\
RAFNet~\cite{wu2022robustrgbd} & 0.883 & 0.901 & 0.838 & 0.074 & 0.891 & 0.966 & 0.897 & 0.055 & 0.826 & 0.858 & 0.830 & 0.118 \\
PopNet~\cite{wu2023popnet} & 0.889 & 0.907 & 0.846 & 0.070 & 0.900 & 0.973 & 0.905 & 0.050 & 0.835 & 0.864 & 0.838 & 0.112 \\
HiDANet~\cite{wu2023hidanet} & 0.876 & 0.894 & 0.832 & 0.078 & 0.883 & 0.962 & 0.892 & 0.059 & 0.819 & 0.853 & 0.823 & 0.125 \\
SwinNet~\cite{liu2022swinnet} & 0.902 & 0.925 & 0.872 & 0.059 & 0.922 & 0.976 & 0.914 & 0.037 & 0.835 & 0.878 & 0.837 & 0.091 \\
VST~\cite{liu2021vst} & 0.890 & 0.914 & 0.858 & 0.065 & 0.912 & 0.968 & 0.902 & 0.044 & 0.820 & 0.865 & 0.823 & 0.101\\
CAVER~\cite{pang2023caver} & 0.899 & 0.911 & 0.853 & 0.071 & 0.906 & \ranksecond{0.979} & 0.913 & 0.054 & \ranksecond{0.842} & 0.870 & 0.844 & 0.113 \\
C$^2$DFNet~\cite{zhang2023c2dfnet} & 0.887 & 0.913 & 0.856 & 0.064 & 0.908 & 0.969 & 0.901 & 0.042 & 0.818 & 0.862 & 0.823 & 0.098 \\
RD3D+~\cite{chen2024rd3d} & 0.893 & 0.918 & 0.864 & 0.061 & 0.915 & 0.971 & 0.907 & 0.039 & 0.825 & 0.870 & 0.831 & 0.095 \\
FasterSal~\cite{zhang2025fastersal} & 0.884 & 0.910 & 0.852 & 0.066 & 0.906 & 0.968 & 0.898 & 0.043 & 0.817 & 0.862 & 0.822 & 0.100 \\
LAFB~\cite{wang2024lafb} & 0.900 & 0.923 & 0.870 & 0.058 & 0.921 & 0.975 & 0.912 & 0.036 & 0.834 & 0.876 & 0.836 & 0.088 \\
PICRNet~\cite{cong2023picrnet} & 0.902 & 0.925 & 0.874 & 0.056 & 0.925 & 0.976 & 0.916 & 0.034 & 0.835 & 0.879 & 0.839 & 0.085 \\
MAGNet~\cite{zhong2024magnet} & 0.904 & 0.927 & 0.878 & 0.054 & 0.927 & 0.977 & 0.920 & 0.032 & 0.836 & 0.881 & 0.842 & 0.083 \\
HFMDNet~\cite{luo2024hfmdnet} & 0.905 & 0.928 & 0.879 & 0.053 & 0.927 & 0.977 & 0.921 & 0.031 & 0.837 & 0.882 & 0.844 & 0.082 \\
EM-Trans~\cite{chen2025emtrans} & 0.903 & 0.927 & 0.876 & 0.054 & 0.925 & 0.976 & 0.918 & 0.033 & 0.835 & 0.880 & 0.841 & 0.084 \\
UniTR~\cite{guo2024unitr} & 0.906 & 0.930 & 0.882 & 0.052 & 0.926 & 0.976 & 0.921 & 0.030 & 0.838 & 0.883 & 0.844 & 0.080 \\
FreqSal~\cite{lyu2026freqsal} & 0.911 & \ranksecond{0.937} & \ranksecond{0.890} & 0.047 & \ranksecond{0.932} & 0.976 & \ranksecond{0.922} & 0.023 & 0.839 & \ranksecond{0.886} & 0.846 & \ranksecond{0.077} \\
\midrule
LSNet~\cite{zhou2023lsnet} & 0.887 & 0.902 & 0.842 & 0.074 & 0.894 & 0.970 & 0.902 & 0.057 & 0.830 & 0.861 & 0.833 & 0.119 \\
LSNet+\method{} (Ours) & \rankbest{0.913} & 0.927 & 0.879 & \ranksecond{0.039} & 0.931 & 0.975 & 0.918 & \ranksecond{0.021} & 0.839 & 0.869 & 0.835 & 0.085 \\
ConTriNet~\cite{tang2025contrinet} & 0.904 & 0.917 & 0.859 & 0.068 & 0.912 & \rankbest{0.982} & 0.917 & 0.052 & \rankbest{0.848} & 0.875 & \ranksecond{0.849} & 0.110 \\
ConTriNet+\method{} (Ours) & \ranksecond{0.912} & \rankbest{0.946} & \rankbest{0.912} & \rankbest{0.034} & \rankbest{0.935} & 0.977 & \rankbest{0.928} & \rankbest{0.016} & \ranksecond{0.842} & \rankbest{0.892} & \rankbest{0.850} & \rankbest{0.074} \\
\bottomrule
\end{tabular*}
}
\WideTable{
{\footnotesize\centering TABLE~\ref{tab:rgbd_sota_full} (continued)\par}\smallskip
\centering\tablefontplain
\begin{tabular*}{\textwidth}{@{\extracolsep{\fill}}lcccccccccccc@{}}
\toprule
Method & \multicolumn{4}{c}{NLPR} & \multicolumn{4}{c}{SIP} & \multicolumn{4}{c}{Mean} \\
\cmidrule(lr){2-5}\cmidrule(lr){6-9}\cmidrule(lr){10-13}
 & $S_m\uparrow$ & adpE$\uparrow$ & adpF$\uparrow$ & MAE$\downarrow$ & $S_m\uparrow$ & adpE$\uparrow$ & adpF$\uparrow$ & MAE$\downarrow$ & $S_m\uparrow$ & adpE$\uparrow$ & adpF$\uparrow$ & MAE$\downarrow$ \\
\midrule
CGFNet~\cite{wang2022cgfnet} & 0.848 & 0.931 & 0.845 & 0.069 & 0.844 & 0.899 & 0.845 & 0.097 & 0.850 & 0.904 & 0.844 & 0.087 \\
TBINet~\cite{wang2022tbinet} & 0.858 & 0.939 & 0.853 & 0.066 & 0.854 & 0.907 & 0.853 & 0.093 & 0.860 & 0.912 & 0.852 & 0.084\\
RAFNet~\cite{wu2022robustrgbd} & 0.862 & 0.941 & 0.857 & 0.063 & 0.858 & 0.910 & 0.857 & 0.089 & 0.864 & 0.915 & 0.856 & 0.080 \\
PopNet~\cite{wu2023popnet} & 0.873 & 0.950 & 0.867 & 0.059 & 0.868 & 0.919 & 0.867 & 0.084 & 0.873 & 0.923 & 0.865 & 0.075 \\
HiDANet~\cite{wu2023hidanet} & 0.855 & 0.937 & 0.851 & 0.067 & 0.851 & 0.905 & 0.851 & 0.094 & 0.857 & 0.910 & 0.850 & 0.085 \\
SwinNet~\cite{liu2022swinnet} & 0.908 & 0.952 & 0.889 & 0.041 & 0.886 & 0.925 & 0.879 & 0.057 & 0.891 & 0.931 & 0.878 & 0.057 \\
VST~\cite{liu2021vst} & 0.896 & 0.944 & 0.878 & 0.047 & 0.871 & 0.914 & 0.866 & 0.064 & 0.878 & 0.921 & 0.865 & 0.064\\
CAVER~\cite{pang2023caver} & 0.878 & 0.954 & 0.872 & 0.061 & 0.874 & 0.922 & 0.872 & 0.086 & 0.880 & 0.927 & 0.871 & 0.077 \\
C$^2$DFNet~\cite{zhang2023c2dfnet} & 0.893 & 0.944 & 0.878 & 0.045 & 0.867 & 0.912 & 0.864 & 0.062 & 0.875 & 0.920 & 0.864 & 0.062 \\
RD3D+~\cite{chen2024rd3d} & 0.899 & 0.947 & 0.883 & 0.043 & 0.878 & 0.918 & 0.872 & 0.060 & 0.882 & 0.925 & 0.871 & 0.060 \\
FasterSal~\cite{zhang2025fastersal} & 0.890 & 0.943 & 0.876 & 0.046 & 0.867 & 0.913 & 0.864 & 0.063 & 0.873 & 0.919 & 0.862 & 0.064 \\
LAFB~\cite{wang2024lafb} & 0.905 & 0.951 & 0.887 & 0.041 & 0.887 & 0.924 & 0.880 & 0.055 & 0.889 & 0.930 & 0.877 & 0.056 \\
PICRNet~\cite{cong2023picrnet} & 0.909 & 0.953 & 0.890 & 0.038 & 0.884 & 0.923 & 0.878 & 0.052 & 0.891 & 0.931 & 0.879 & 0.053 \\
MAGNet~\cite{zhong2024magnet} & 0.911 & 0.954 & 0.893 & 0.036 & 0.891 & 0.927 & 0.883 & 0.049 & 0.894 & 0.933 & 0.883 & 0.051 \\
HFMDNet~\cite{luo2024hfmdnet} & 0.910 & 0.954 & 0.892 & 0.035 & 0.885 & 0.924 & 0.880 & 0.050 & 0.893 & 0.933 & 0.883 & 0.050 \\
EM-Trans~\cite{chen2025emtrans} & 0.908 & 0.953 & 0.891 & 0.036 & 0.884 & 0.925 & 0.880 & 0.051 & 0.891 & 0.932 & 0.881 & 0.052 \\
UniTR~\cite{guo2024unitr} & 0.909 & 0.953 & 0.893 & 0.034 & 0.887 & 0.932 & 0.886 & 0.048 & 0.893 & 0.935 & 0.885 & 0.049 \\
FreqSal~\cite{lyu2026freqsal} & \ranksecond{0.916} & 0.954 & \ranksecond{0.894} & \ranksecond{0.027} & \ranksecond{0.894} & \ranksecond{0.935} & \ranksecond{0.889} & \ranksecond{0.041} & \ranksecond{0.898} & \ranksecond{0.938} & \ranksecond{0.888} & \ranksecond{0.043} \\
\midrule
LSNet~\cite{zhou2023lsnet} & 0.866 & 0.945 & 0.861 & 0.064 & 0.862 & 0.913 & 0.861 & 0.090 & 0.868 & 0.918 & 0.860 & 0.081 \\
LSNet+\method{} (Ours) & \rankbest{0.919} & \ranksecond{0.955} & 0.887 & \ranksecond{0.027} & 0.878 & 0.916 & 0.872 & 0.055 & 0.896 & 0.928 & 0.878 & 0.045 \\
ConTriNet~\cite{tang2025contrinet} & 0.883 & \rankbest{0.959} & 0.878 & 0.058 & 0.879 & 0.927 & 0.878 & 0.083 & 0.885 & 0.932 & 0.876 & 0.074 \\
ConTriNet+\method{} (Ours) & 0.915 & \ranksecond{0.955} & \rankbest{0.896} & \rankbest{0.023} & \rankbest{0.898} & \rankbest{0.940} & \rankbest{0.903} & \rankbest{0.038} & \rankbest{0.901} & \rankbest{0.942} & \rankbest{0.898} & \rankbest{0.037} \\
\bottomrule
\end{tabular*}
}

\WideTable{
\caption{\focus{Transfer to lightweight DFM-Net and transformer-based VST-T++ on five RGB-D benchmarks.} Both rows of each detector are trained and evaluated under the same protocol (Sec.~4.1); values are five-seed means. Each benchmark cell reports $S_m$/MAE.}
\label{tab:vstpp_transfer}
\centering
\tablefontdense
\begin{tabular*}{\linewidth}{@{\extracolsep{\fill}}llcccccc@{}}
\toprule
Detector & Configuration & NJU2K & DES & LFSD & NLPR & SIP & $\Delta$P (M) \\
\midrule
\multirow{2}{*}{DFM-Net~\cite{zhang2021dqfm}} & Baseline & 0.861/0.045 & 0.884/0.022 & 0.822/0.077 & 0.877/0.028 & 0.839/0.055 & 0 \\
 & +MD--FMRR & \focus{0.875/0.041} & \focus{0.898/0.020} & \focus{0.837/0.070} & \focus{0.891/0.025} & \focus{0.854/0.049} & \focus{\VOneDFMParamOverhead} \\
\midrule
\multirow{2}{*}{VST-T++~\cite{liu2024vstpp}} & Baseline & 0.893/0.026 & 0.902/0.015 & 0.844/0.065 & 0.891/0.022 & 0.866/0.038 & 0 \\
 & +MD--FMRR & \focus{0.907/0.024} & \focus{0.916/0.014} & \focus{0.859/0.059} & \focus{0.905/0.020} & \focus{0.880/0.034} & \focus{\VOneLSParamOverhead} \\
\bottomrule
\end{tabular*}
}

\rev{\focus{Each FMRR-enhanced model is paired with its own reproduced baseline under matched backbone-specific training schedules. The comparison therefore evaluates the complete MD--FMRR configuration within each detector. Rows labeled ``+FMRR (Ours)'' identify our method; the adjacent unmodified detector rows are its matched baselines. DFM-Net and VST-T++ extend this evaluation to dedicated RGB-D architectures with different interaction designs, while retaining their original fusion and decoding pathways. Together, these integrations test the same participation rule across both sensing combinations and backbone families.}}

\subsection{\rev{Emergence from Demand and Pathway}}
\label{sec:factorial}
\rev{\focus{Table~\ref{tab:factorial} tests the demand--pathway mechanism directly. A1 introduces availability contrast without a participation path, and A2 provides the path without repeated availability contrast. A3 is the only setting that combines both elements and consistently yields strong complete-, RGB-only-, and auxiliary-only prediction on both modality combinations.}}

\rev{\focus{A1 exposes the detector to complete and single-modality states without adding a pathway that can change how much each stream participates in fusion. Under the stabilized A1 setting, complete-input $S_m$ is 0.010 below A0 while both single-modality states improve in all four primary task--backbone combinations. Related availability trade-offs have been reported for dual-modal SOD~\cite{hao2024cola} and audio-visual recognition~\cite{dai2024modalitybias}. Tables~\ref{tab:factorial} and~\ref{tab:dedicated_rgbd_matched} show the same pattern across ConTriNet, LSNet, DFM-Net, and VST-T++. A3 retains the original backbone learning rate and exceeds A0 in every input state.}}

\WideBlocks{
\begin{columntable}
\caption{Demand--pathway factorial analysis. \focus{A0 is native, A1 adds MD, A2 adds FMRR, and A3 combines both.} Values are five-run test-set $S_m$ task means; \focus{A1 reports converged runs at stabilized rates (Sec.~4.3).}}
\label{tab:factorial}
\centering
\tablefont\renewcommand{\arraystretch}{1.12}\setlength{\tabcolsep}{2.4pt}
\begin{minipage}[t]{\linewidth}
\vspace{0pt}\centering
\begin{tabular*}{\linewidth}{@{\extracolsep{\fill}}lccc@{}}
\toprule
Setting & Full & RGB only & Auxiliary only \\
\midrule
\multicolumn{4}{l}{\textbf{RGB-D / ConTriNet}} \\
A0 & $0.885$ & $0.855$ & $0.736$ \\
A1 & \focus{$0.875$} & \focus{$0.870$} & \focus{$0.831$} \\
A2 & $0.896$ & $0.861$ & $0.755$ \\
A3 & \focus{\textbf{$0.901$}} & \focus{\textbf{$0.893$}} & \focus{\textbf{$0.890$}} \\
\midrule
\multicolumn{4}{l}{\textbf{RGB-D / LSNet}} \\
A0 & $0.868$ & $0.839$ & $0.703$ \\
A1 & \focus{$0.858$} & \focus{$0.846$} & \focus{$0.742$} \\
A2 & $0.882$ & $0.846$ & $0.728$ \\
A3 & \focus{\textbf{$0.896$}} & \focus{\textbf{$0.856$}} & \focus{\textbf{$0.765$}} \\
\bottomrule
\end{tabular*}
\end{minipage}\par\smallskip
\begin{minipage}[t]{\linewidth}
\vspace{0pt}\centering
\begin{tabular*}{\linewidth}{@{\extracolsep{\fill}}lccc@{}}
\toprule
Setting & Full & RGB only & Auxiliary only \\
\midrule
\multicolumn{4}{l}{\textbf{RGB-T / ConTriNet}} \\
A0 & $0.905$ & \vresult{$0.706$} & \vresult{$0.840$} \\
A1 & \focus{$0.895$} & \focus{\vresult{$0.793$}} & \focus{\vresult{$0.871$}} \\
A2 & $0.914$ & \vresult{$0.739$} & \vresult{$0.852$} \\
A3 & \focus{\textbf{$0.924$}} & \focus{\textbf{\vresult{$0.905$}}} & \focus{\textbf{\vresult{$0.894$}}} \\
\midrule
\multicolumn{4}{l}{\textbf{RGB-T / LSNet}} \\
A0 & $0.892$ & \vresult{$0.698$} & \vresult{$0.837$} \\
A1 & \focus{$0.882$} & \focus{\vresult{$0.761$}} & \focus{\vresult{$0.857$}} \\
A2 & $0.893$ & \vresult{$0.721$} & \vresult{$0.844$} \\
A3 & \focus{\textbf{$0.894$}} & \focus{\textbf{\vresult{$0.859$}}} & \focus{\textbf{\vresult{$0.862$}}} \\
\bottomrule
\end{tabular*}
\end{minipage}
\end{columntable}
}

\WideBlocks{
\TopColumns{}{

}{
\subsection{\rev{Matched Admission-Pathway Comparison}}
\label{sec:admission_comparison}
\rev{\focus{Table~\ref{tab:admission} compares participation pathways with MD, data splits, schedules, and downstream detectors held fixed. The availability-flag hard gate is strongest on exactly missing inputs, where the flag identifies the unavailable stream. On nonzero and graded inputs the flag no longer separates useful from harmful evidence, and FMRR gives the best results. The foreground-area and background-response statistics in Appendix~C compare A0 with the complete MD--FMRR configuration, showing how the combined strategy recovers foreground extent and suppresses background responses.}}

\begin{columntable}
\caption{\focus{Comparison of pre-fusion admission mechanisms under matched MD training. Scores are five-seed means averaged over ConTriNet and LSNet on RGB-D and RGB-T. ``Nonzero'' averages constant, mean, low-variance, and saturated substitutes; ``graded'' averages the held-out severity scans. The availability-flag hard gate closes a stream only when it is flagged as missing; nonzero substitutes and graded degradations are flagged as available.}}
\label{tab:admission}
\centering
\tablefontdense\renewcommand{\arraystretch}{1.25}\setlength{\tabcolsep}{0.7pt}
\begin{tabular*}{\linewidth}{@{\extracolsep{\fill}}P{0.38\linewidth}cccccc@{}}
\toprule
Admission pathway & \multicolumn{1}{c}{Full} & \multicolumn{1}{c}{RGB} & \multicolumn{1}{c}{Aux.} & \multicolumn{1}{c}{Nonzero} & \multicolumn{1}{c}{Graded} & \multicolumn{1}{c}{$F_\beta^w$} \\
\midrule
Availability-flag hard gate & 0.899 & \focus{\textbf{0.883}} & \focus{\textbf{0.863}} & 0.824 & \vresult{0.850} & 0.877 \\
Missing token & 0.894 & 0.871 & 0.848 & 0.831 & \vresult{0.840} & 0.866 \\
Avail.-cond. FiLM & 0.896 & 0.873 & 0.851 & 0.833 & \vresult{0.846} & 0.871 \\
Conditional dropout & 0.898 & 0.875 & 0.853 & 0.838 & \vresult{0.855} & 0.878 \\
Input-norm gate & 0.891 & 0.868 & 0.844 & 0.815 & \vresult{0.831} & 0.859 \\
FMRR & \focus{\textbf{0.904}} & \focus{\vresult{0.878}} & \focus{\vresult{0.853}} & \focus{\textbf{0.845}} & \focus{\textbf{\vresult{0.872}}} & \focus{\textbf{0.886}} \\
\bottomrule
\end{tabular*}
\end{columntable}

\subsection{\rev{Participation Tracks External Modality Value}}
\label{sec:external_value}
\rev{\focus{To relate participation to predictive evidence, Table~\ref{tab:external_value} compares $w^r-w^a$ with the per-sample advantage of independently trained RGB-only and auxiliary-only detectors. Positive rank agreement is maintained across complete, missing, nonzero, local, graded, and real sensor states. The pooled RGB-D and RGB-T correlations are 0.80 and 0.76, respectively, while the severity ordering reaches Kendall $\tau=0.80$.}}

\rev{\focus{Fig.~\ref{fig:severity_scan} provides the graded view: as evidence degrades, the affected stream loses participation and the routed detector separates progressively from blind fusion. Fig.~\ref{fig:share_distributions} provides the sample distribution: thermal evidence is favored in low illumination, RGB is favored during thermal crossover and depth holes, and the complete-input distribution remains centered near equal participation. Appendix~E reports relative-misalignment robustness and the complementary photometric-stability test.}}

\begin{columntable}
\caption{\focus{Agreement between learned relative participation $w^r-w^a$ and external modality-value references. $\rho$ compares the share difference with the per-sample $S_m$ advantage of independently trained RGB-only and auxiliary-only detectors; $\tau$ measures monotonic response to degradation severity. Regret is the mean $S_m$ gap to the better single-modality reference, in $10^{-3}$.}}
\label{tab:external_value}
\centering
\tablefontdense\renewcommand{\arraystretch}{1.25}
\begin{tabular*}{\linewidth}{@{\extracolsep{\fill}}l*{4}{c}@{}}
\toprule
State group & \multicolumn{1}{c}{\shortstack{RGB-D\\$\rho\uparrow$}} & \multicolumn{1}{c}{\shortstack{RGB-T\\$\rho\uparrow$}} & \multicolumn{1}{c}{\shortstack{Severity\\$\tau\uparrow$}} & \multicolumn{1}{c}{Regret $\downarrow$} \\
\midrule
Complete inputs & 0.74 & 0.70 & -- & 3.2 \\
Zero-valued intervention & \textbf{0.91} & \textbf{0.90} & -- & \textbf{0.8} \\
Nonzero substitutes & 0.81 & 0.78 & 0.76 & 1.8 \\
Local corruptions & 0.77 & 0.73 & 0.71 & 2.4 \\
Degradation severity & \vresult{0.82} & \vresult{0.79} & \vresult{0.85} & \vresult{1.8} \\
Real sensor attributes & 0.79 & 0.75 & -- & 2.1 \\
\midrule
Pooled & \focus{\textbf{0.80}} & \focus{\textbf{0.76}} & \focus{\textbf{\vresult{0.80}}} & \focus{\textbf{2.0}} \\
\bottomrule
\end{tabular*}
\end{columntable}

\begin{columnfigure}
\centering
\includegraphics[width=\linewidth]{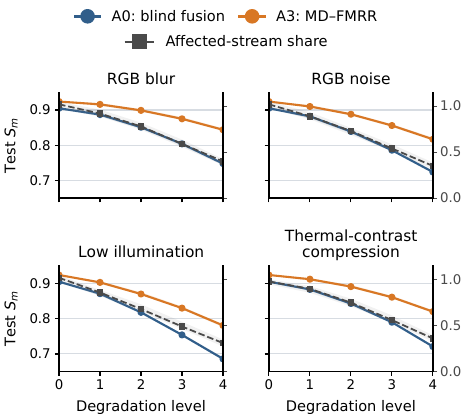}
\caption{\rev{Five-level RGB-T degradation results for ConTriNet. MD--FMRR retains higher $S_m$ as the affected-stream share decreases. Shading spans the five runs.}}
\label{fig:severity_scan}
\end{columnfigure}

\begin{columnfigure}
\centering
\includegraphics[trim=5bp 8bp 5bp 5bp,clip,width=\linewidth]{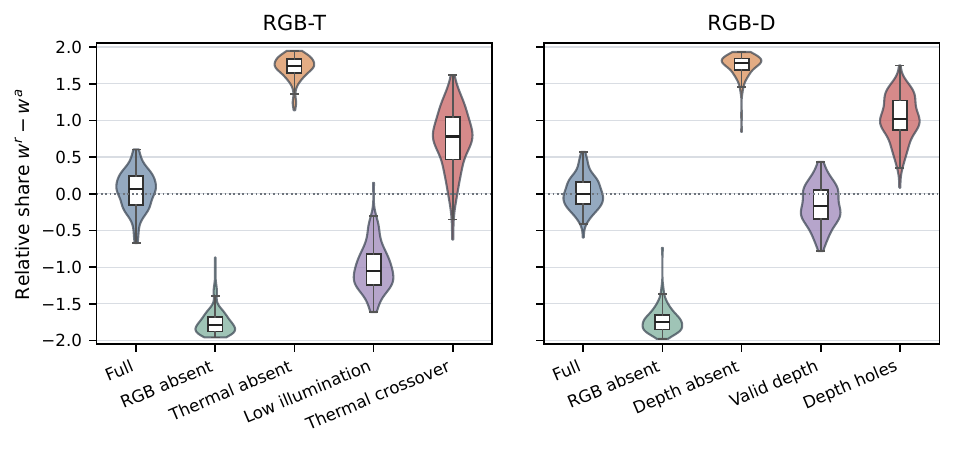}
\caption{\rev{Distribution of relative participation across complete, unavailable, and naturally challenging sensor states. Positive values favor RGB and negative values favor the auxiliary stream.}}
\label{fig:share_distributions}
\end{columnfigure}
\subsection{\rev{Transfer to Unseen Modality States}}
\label{sec:unseen_states}
\rev{\focus{Table~\ref{tab:unseen_missing} evaluates states that are not the zero-valued intervention used by the main training configuration. FMRR improves over blind fusion and the input-norm gate on nonzero substitutes and on every local-mask extent. Consistent with Eq.~\eqref{eq:input_normalization}, spatially constant all-zero and all-one auxiliary inputs produce the same normalized router observation and receive the same depth share; their final predictions differ because the original inputs still generate different encoder features. For rectangular-mask coverages of 25\%, 50\%, and 75\%, the mean depth shares are 0.82, 0.59, and 0.36, respectively, linking allocation to the remaining evidence in the affected modality.}}
}
\begin{columntable}
\caption{\focus{Generalization from zero-valued modality dropout to nonzero and spatially partial representations of the affected depth stream. Performance cells report five-seed mean $S_m$ for ConTriNet on the mean RGB-D test protocol; the final column reports the corresponding mean FMRR depth share. Under Eq.~(5), spatially constant zero and one inputs produce the same normalized router observation and therefore the same depth share. Oracle drop removes the affected depth stream using the known perturbation identity and serves as a reference. Bold marks the best learned routing method.}}
\label{tab:unseen_missing}
\centering
\tablefontdense\renewcommand{\arraystretch}{1.16}
\begin{tabular*}{\linewidth}{@{\extracolsep{\fill}}P{.28\linewidth}ccccc@{}}
\toprule
\shortstack[l]{Affected-input\\representation} & \multicolumn{1}{c}{\shortstack{Blind\\fusion}} & \multicolumn{1}{c}{\shortstack{Input-norm\\gate}} & \multicolumn{1}{c}{MD--FMRR} & \multicolumn{1}{c}{\shortstack{Oracle\\drop}} & \multicolumn{1}{c}{\shortstack{Depth\\share}} \\
\midrule
All zero & 0.855 & 0.884 & \focus{\textbf{0.893}} & 0.895 & \focus{0.15} \\
All one & 0.652 & 0.655 & \focus{\textbf{0.842}} & 0.895 & \focus{\vresult{0.15}} \\
Dataset-mean image & 0.671 & 0.676 & \focus{\textbf{0.851}} & 0.895 & 0.34 \\
Low-variance Gaussian field & 0.681 & 0.687 & \focus{\textbf{0.836}} & 0.895 & 0.26 \\
Saturated image & 0.648 & 0.659 & \focus{\textbf{0.821}} & 0.895 & 0.21 \\
25\% rectangular mask & 0.842 & 0.843 & \focus{\textbf{0.882}} & 0.895 & \focus{0.82} \\
50\% rectangular mask & 0.797 & 0.799 & \focus{\textbf{0.865}} & 0.895 & \focus{0.59} \\
75\% rectangular mask & 0.748 & 0.764 & \focus{\textbf{0.841}} & 0.895 & \focus{0.36} \\
Irregular local holes & 0.805 & 0.806 & \focus{\textbf{0.871}} & 0.895 & 0.52 \\
\bottomrule
\end{tabular*}
\end{columntable}

}

\WideBlocks{
\AlignedColumns{

\subsection{Qualitative Results}
\label{loc:qualitative}
\rev{\focus{Fig.~\ref{fig:qualitative} shows complete and single-modality predictions. Appendix~H discusses extreme weather and motivates finer spatial routing.}}

}{
\TopFigure{
\centering
\includegraphics[trim=11bp 27bp 0bp 14bp,clip,width=0.88\textwidth]{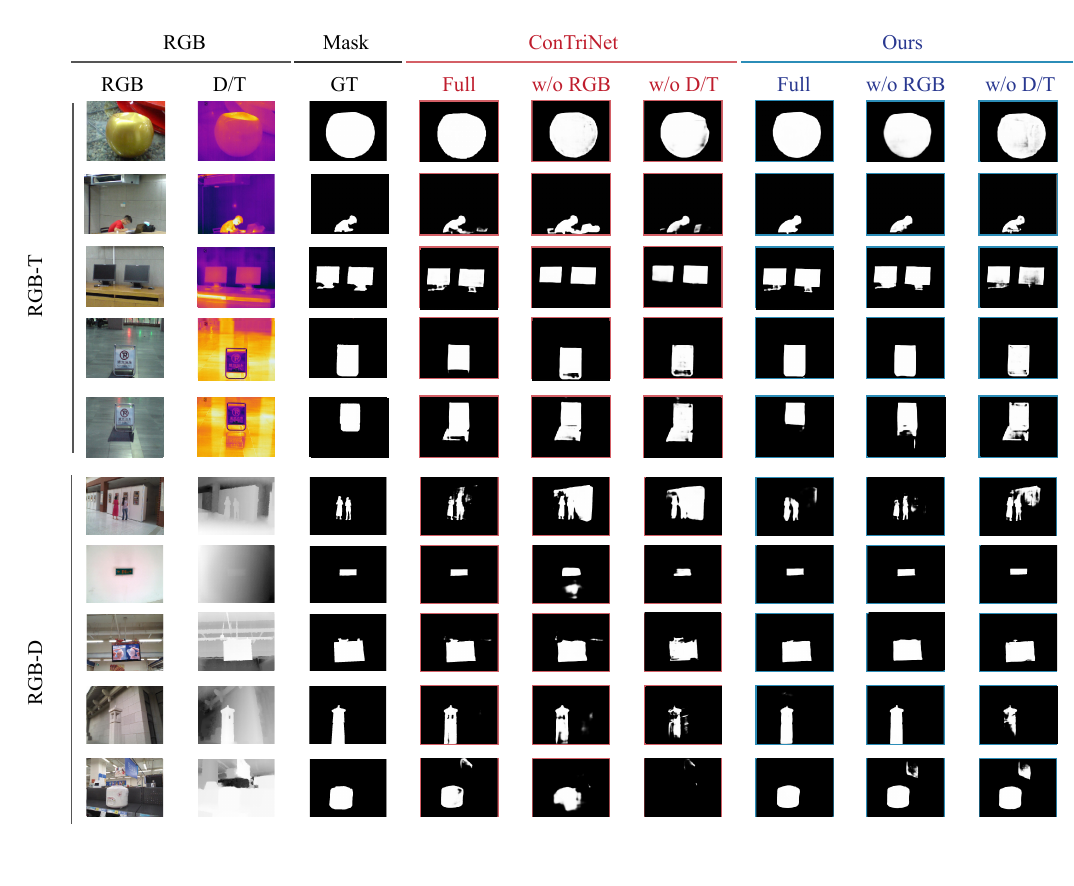}
\caption{\rev{Qualitative comparison under complete and single-modality inputs. D/T denotes the depth or thermal stream. The routed model preserves the salient structure as the available evidence changes.}}
\label{fig:qualitative}
}

\begin{columntable}
\caption{\focus{Global-state estimators for ConTriNet on the mean RGB-D test protocol, with the same allocator and MD schedule. Scores are five-seed mean full/drop/unseen $S_m$: drop averages both single-modality states; unseen averages nonzero and local conditions in Table~\ref{tab:unseen_missing}. Extra parameters are in millions; $\rho$ and regret ($10^{-3}$) follow Table~\ref{tab:external_value}.}}
\label{tab:estimator}
\centering
\tablefontdense
\begin{tabular*}{\linewidth}{@{\extracolsep{\fill}}P{0.25\linewidth}cccc@{}}
\toprule
Estimator & \shortstack{Extra params\\(M)} & Scores & \multicolumn{1}{c}{$\rho$} & \multicolumn{1}{c}{Regret} \\
\midrule
Spatial statistics & 0.001 & 0.889/0.845/0.812 & 0.51 & 4.7 \\
Quality vector & 0.003 & 0.892/0.855/0.825 & 0.60 & 3.9 \\
Matched spatial CNN & \focus{0.247} & \focus{0.897/0.883/0.844} & \focus{0.69} & \focus{2.9} \\
Unfiltered FFT & 0.001 & 0.895/0.879/0.838 & 0.66 & 3.1 \\
FMRR & \focus{0.249} & \focus{\textbf{0.901/0.892/0.851}} & \focus{\textbf{0.80}} & \focus{\textbf{2.0}} \\
\bottomrule
\end{tabular*}
\end{columntable}

\section{Ablation Study}
\subsection{\rev{Global-State Estimator}}
\label{sec:estimator_ablation}
\rev{\focus{Table~\ref{tab:estimator} keeps the participation interface fixed and changes only the observation space. Filtered Fourier statistics provide the best combined complete-input, missing-input, unseen-state, and external-value result. The parameter-matched spatial CNN is the closest alternative. Against this model, the gains are larger for missing and unseen inputs than for complete inputs. The unfiltered FFT row separates the value of global amplitude statistics from the learnable filter.}}

\subsection{\rev{Participation-Pathway Design}}
\label{loc:pathway}
\rev{Table~\ref{tab:routing_design} compares designs within the same participation interface. The selected configuration combines per-image normalization, mean/\allowbreak std/\allowbreak max spectral statistics, a learnable filter, joint observation, fixed-budget softmax, a 16-dimensional hidden layer, and hierarchy-wide insertion. These choices improve share--value agreement and unseen-state prediction with 0.249\,M added parameters.}
}
\WideTable{
\caption{\focus{Participation-pathway ablation for ConTriNet on the mean RGB-D test protocol. Scores are five-seed means reported as full/drop/unseen $S_m$; every block changes one factor from the selected configuration, marked in bold.}}
\label{tab:routing_design}
\centering
\tablefont\setlength{\tabcolsep}{2.2pt}\renewcommand{\arraystretch}{1.08}
\begin{minipage}[t]{0.485\linewidth}
\vspace{0pt}\centering
\begin{tabular*}{\linewidth}{@{\extracolsep{\fill}}lccc@{}}
\toprule
Variant & \shortstack{Extra params\\(M)} & Scores & $\rho$ \\
\midrule
\multicolumn{4}{l}{\textbf{Statistics}} \\
Mean only & 0.249 & 0.895/0.881/0.840 & 0.67 \\
Mean + std & 0.249 & 0.898/0.887/0.846 & 0.74 \\
\textbf{Mean + std + max} & 0.249 & \textbf{0.901/0.892/0.851} & \textbf{0.80} \\
\midrule
\multicolumn{4}{l}{\textbf{Spectral filter}} \\
Fixed identity & 0.001 & 0.895/0.879/0.838 & 0.66 \\
\textbf{Learnable} & 0.249 & \textbf{0.901/0.892/0.851} & \textbf{0.80} \\
\midrule
\multicolumn{4}{l}{\textbf{Input normalization}} \\
None & 0.249 & 0.896/0.883/0.842 & 0.68 \\
\textbf{Per-image/channel} & 0.249 & \focus{\textbf{0.901/0.892/0.851}} & \focus{\textbf{0.80}} \\
\midrule
\multicolumn{4}{l}{\textbf{Allocation}} \\
Independent predictors & 0.249 & 0.897/0.885/0.843 & 0.70 \\
\textbf{Joint descriptor} & 0.249 & \focus{\textbf{0.901/0.892/0.851}} & \focus{\textbf{0.80}} \\
\bottomrule
\end{tabular*}
\end{minipage}\hfill
\begin{minipage}[t]{0.485\linewidth}
\vspace{0pt}\centering\renewcommand{\arraystretch}{1.11}
\begin{tabular*}{\linewidth}{@{\extracolsep{\fill}}lccc@{}}
\toprule
Variant & \shortstack{Extra params\\(M)} & Scores & $\rho$ \\
\midrule
\multicolumn{4}{l}{\textbf{Output constraint}} \\
Independent $[0,1]$ & 0.249 & 0.895/0.878/0.838 & 0.65 \\
Independent $[0,2]$ & 0.249 & 0.898/0.886/0.844 & 0.72 \\
\textbf{$2\!\cdot\!$softmax} & 0.249 & \focus{\textbf{0.901/0.892/0.851}} & \focus{\textbf{0.80}} \\
\midrule
\multicolumn{4}{l}{\textbf{Hidden dimension}} \\
8 & 0.249 & 0.898/0.887/0.846 & 0.75 \\
\textbf{16} & 0.249 & \textbf{0.901/0.892/0.851} & \textbf{0.80} \\
32 & 0.250 & 0.900/0.890/0.849 & 0.79 \\
\midrule
\multicolumn{4}{l}{\textbf{Insertion position}} \\
Raw input & 0.249 & 0.894/0.879/0.839 & 0.68 \\
Shallow feature & 0.249 & 0.897/0.885/0.845 & 0.73 \\
Deep feature & 0.249 & 0.896/0.883/0.843 & 0.71 \\
\textbf{Complete hierarchy} & 0.249 & \textbf{0.901/0.892/0.851} & \textbf{0.80} \\
\bottomrule
\end{tabular*}
\end{minipage}
}

}

\WideBlocks{
\WideIntro{\rmfamily\normalsize
\rev{Table~\ref{tab:severity_absolute} gives absolute degradation metrics. Supplementary Tables~\ref{tab:rgbt_robustness_mean}--\ref{tab:rgbd_robustness_mean} report missing-input results; Table~\ref{tab:single_modality_stats} characterizes foreground recovery and background suppression. Tables~\ref{tab:integration_details}--\ref{tab:dedicated_rgbd_matched} detail integration and transfer, and Table~\ref{tab:factorial_seeds} lists individual runs. Tables~\ref{tab:stability}--\ref{tab:md_forms} examine observation stability and training settings. Table~\ref{tab:attribute_results} covers natural sensor conditions, while Table~\ref{tab:single_reference} specifies the independent modality-value references.}
}
\WideTable{
\caption{ \focus{Absolute results underlying the five-level trajectories in Fig.~\ref{fig:severity_scan}}, for ConTriNet on the RGB-T test protocol (mean over VT821, VT1000, and VT5000). The clean row is shared by all four families and matches Table~\ref{tab:rgbt_sota_full}. Values are five-seed means; higher is better except for MAE.}\label{tab:severity_absolute}
\sffamily\tablefontdense\renewcommand{\arraystretch}{1.02}
\begin{tabular*}{\textwidth}{@{\extracolsep{\fill}}lc*{10}{c}@{}}
\toprule
\multirow{2}{*}{Condition} & \multirow{2}{*}{Level} & \multicolumn{5}{c}{A0: blind fusion} & \multicolumn{5}{c}{A3: MD--FMRR} \\
\cmidrule(lr){3-7}\cmidrule(lr){8-12}
 & & $S_m$ & MAE & adpE & adpF & $F_\beta^w$ & $S_m$ & MAE & adpE & adpF & $F_\beta^w$ \\
\midrule
Clean & 0 & \vresult{0.905} & \vresult{0.029} & \vresult{0.928} & \vresult{0.860} & \vresult{0.858} & \vresult{0.924} & \vresult{0.024} & \vresult{0.943} & \vresult{0.893} & \vresult{0.891} \\
RGB blur & 1 & \vresult{0.887} & \vresult{0.038} & \vresult{0.919} & \vresult{0.846} & \vresult{0.844} & \vresult{0.916} & \vresult{0.029} & \vresult{0.939} & \vresult{0.887} & \vresult{0.885} \\\*
 & 2 & \vresult{0.851} & \vresult{0.052} & \vresult{0.900} & \vresult{0.823} & \vresult{0.821} & \vresult{0.899} & \vresult{0.037} & \vresult{0.929} & \vresult{0.876} & \vresult{0.874} \\\*
 & 3 & \vresult{0.804} & \vresult{0.072} & \vresult{0.873} & \vresult{0.790} & \vresult{0.788} & \vresult{0.875} & \vresult{0.050} & \vresult{0.914} & \vresult{0.860} & \vresult{0.858} \\\*
 & 4 & \vresult{0.748} & \vresult{0.099} & \vresult{0.838} & \vresult{0.751} & \vresult{0.748} & \vresult{0.844} & \vresult{0.068} & \vresult{0.894} & \vresult{0.841} & \vresult{0.838} \\
\midrule
RGB noise & 1 & \vresult{0.882} & \vresult{0.041} & \vresult{0.916} & \vresult{0.842} & \vresult{0.840} & \vresult{0.910} & \vresult{0.031} & \vresult{0.936} & \vresult{0.883} & \vresult{0.881} \\\*
 & 2 & \vresult{0.839} & \vresult{0.059} & \vresult{0.892} & \vresult{0.811} & \vresult{0.809} & \vresult{0.888} & \vresult{0.041} & \vresult{0.924} & \vresult{0.869} & \vresult{0.867} \\\*
 & 3 & \vresult{0.786} & \vresult{0.084} & \vresult{0.858} & \vresult{0.770} & \vresult{0.768} & \vresult{0.856} & \vresult{0.056} & \vresult{0.906} & \vresult{0.849} & \vresult{0.847} \\\*
 & 4 & \vresult{0.724} & \vresult{0.117} & \vresult{0.812} & \vresult{0.714} & \vresult{0.711} & \vresult{0.817} & \vresult{0.078} & \vresult{0.881} & \vresult{0.821} & \vresult{0.818} \\
\midrule
Low illumination & 1 & \vresult{0.871} & \vresult{0.045} & \vresult{0.909} & \vresult{0.832} & \vresult{0.830} & \vresult{0.903} & \vresult{0.033} & \vresult{0.932} & \vresult{0.878} & \vresult{0.876} \\\*
 & 2 & \vresult{0.818} & \vresult{0.067} & \vresult{0.880} & \vresult{0.794} & \vresult{0.791} & \vresult{0.870} & \vresult{0.046} & \vresult{0.915} & \vresult{0.857} & \vresult{0.855} \\\*
 & 3 & \vresult{0.754} & \vresult{0.097} & \vresult{0.840} & \vresult{0.744} & \vresult{0.741} & \vresult{0.830} & \vresult{0.064} & \vresult{0.892} & \vresult{0.828} & \vresult{0.826} \\\*
 & 4 & \vresult{0.686} & \vresult{0.134} & \vresult{0.786} & \vresult{0.680} & \vresult{0.677} & \vresult{0.781} & \vresult{0.089} & \vresult{0.859} & \vresult{0.790} & \vresult{0.787} \\
\midrule
Thermal-contrast compression & 1 & \vresult{0.883} & \vresult{0.040} & \vresult{0.917} & \vresult{0.843} & \vresult{0.841} & \vresult{0.912} & \vresult{0.030} & \vresult{0.937} & \vresult{0.885} & \vresult{0.883} \\\*
 & 2 & \vresult{0.843} & \vresult{0.057} & \vresult{0.895} & \vresult{0.814} & \vresult{0.812} & \vresult{0.891} & \vresult{0.039} & \vresult{0.926} & \vresult{0.872} & \vresult{0.870} \\\*
 & 3 & \vresult{0.790} & \vresult{0.082} & \vresult{0.861} & \vresult{0.773} & \vresult{0.771} & \vresult{0.861} & \vresult{0.054} & \vresult{0.909} & \vresult{0.853} & \vresult{0.851} \\\*
 & 4 & \vresult{0.721} & \vresult{0.119} & \vresult{0.808} & \vresult{0.708} & \vresult{0.705} & \vresult{0.820} & \vresult{0.076} & \vresult{0.884} & \vresult{0.824} & \vresult{0.821} \\
\bottomrule
\end{tabular*}
}

}

\section{Conclusion}
\label{loc:conclusion}
\rev{FMRR couples frequency-domain routing with modality dropout to suppress interference in RGB-D and RGB-T SOD. Joint normalized spectral descriptors determine fixed-budget relative weights that scale complete feature hierarchies before native fusion. The original SOD objective learns these weights through their effects on prediction. Matched comparisons and independent single-modality references link the weights to useful modality evidence. Results on ConTriNet, LSNet, DFM-Net, and VST-T++ show improved complete-input prediction and single-modality fallback. Future work will extend the global weights to spatially finer routing under challenging conditions.}

% Appendices retain two-column prose and use full-width blocks where needed.

\appendices
\section*{Appendices}
\renewcommand{\thesection}{\Alph{section}}
\section{\rev{Degradation Protocol}}
\label{loc:degradation}
\rev{\focus{Every held-out degradation uses five ordered levels and deterministic sample-index seeds. RGB blur uses Gaussian standard deviations of 0, 1.0, 2.0, 3.2, and 4.8; RGB noise uses intensity deviations of 0, 8, 16, 22, and 30. Low illumination uses brightness factors of 1.00, 0.82, 0.64, 0.46, and 0.30. The corresponding gamma values are 1.00, 1.15, 1.35, 1.60, and 1.90. Auxiliary degradation uses thermal-contrast factors of 1.00, 0.82, 0.64, 0.46, and 0.28 for RGB-T, and depth quantization levels of 256, 64, 32, 16, and 8 for RGB-D. Rectangular and irregular holes are sampled independently of ground-truth saliency and then held fixed for every compared model. No held-out degradation enters training or checkpoint selection.}}\par\medskip
\section{\rev{Missing-Input Evaluation}}
\label{loc:missing}
\rev{Tables~\ref{tab:rgbt_robustness_mean} and~\ref{tab:rgbd_robustness_mean} evaluate the same detector checkpoints after one input stream is removed. Each FMRR-enhanced model uses the MD--FMRR configuration described in Sec.~4.3. In the RGB-T results, both routed detectors improve $S_m$, adpE, and adpF while lowering MAE for either removed stream. Appendix~C examines the associated foreground recovery and background response.}

\WideBlocks{
\WideTable{
\caption{RGB-T performance when one stream is removed at inference. \focus{All detectors are trained under the same task protocol; rows labeled ``+FMRR (Ours)'' add the complete MD--FMRR configuration.} Columns report five-seed means.}
\label{tab:rgbt_robustness_mean}
\centering
\tablefontplain
\begin{tabular*}{\textwidth}{@{\extracolsep{\fill}}lcccccccc@{}}
\toprule
Method & \multicolumn{4}{c}{Drop RGB} & \multicolumn{4}{c}{Drop Thermal} \\
\cmidrule(lr){2-5}\cmidrule(lr){6-9}
 & $S_m\uparrow$ & adpE$\uparrow$ & adpF$\uparrow$ & MAE$\downarrow$ & $S_m\uparrow$ & adpE$\uparrow$ & adpF$\uparrow$ & MAE$\downarrow$ \\
\midrule
CGFNet~\cite{wang2022cgfnet} & 0.810 & 0.861 & 0.808 & \vresult{0.077} & 0.676 & 0.724 & 0.679 & \vresult{0.140} \\
TBINet~\cite{wang2022tbinet} & 0.827 & 0.882 & 0.812 & \vresult{0.069} & 0.702 & 0.748 & 0.688 & \vresult{0.129} \\
HiDANet~\cite{wu2023hidanet} & 0.819 & 0.874 & 0.809 & \vresult{0.073} & 0.694 & 0.742 & 0.684 & \vresult{0.133} \\
CAVER~\cite{pang2023caver} & 0.831 & 0.874 & 0.806 & \vresult{0.067} & 0.695 & 0.737 & 0.675 & \vresult{0.132} \\
\midrule
LSNet~\cite{zhou2023lsnet} & 0.837 & 0.864 & 0.794 & \vresult{0.064} & 0.698 & 0.720 & 0.663 & \vresult{0.130} \\
LSNet+\method{} (Ours) & \ranksecond{0.862} & \rankbest{0.935} & 0.811 & \ranksecond{0.051} & \ranksecond{0.859} & \ranksecond{0.897} & \ranksecond{0.781} & \ranksecond{0.043} \\
ConTriNet~\cite{tang2025contrinet} & 0.840 & 0.885 & \ranksecond{0.816} & \vresult{0.061} & 0.706 & 0.750 & 0.687 & \vresult{0.126} \\
ConTriNet+\method{} (Ours) & \rankbest{0.894} & \ranksecond{0.904} & \rankbest{0.853} & \rankbest{0.035} & \rankbest{0.905} & \rankbest{0.924} & \rankbest{0.900} & \rankbest{0.033} \\
\bottomrule
\end{tabular*}
}

\WideTable{
\caption{RGB-D performance when one stream is removed at inference. \focus{All detectors are trained under the same task protocol; rows labeled ``+FMRR (Ours)'' add the complete MD--FMRR configuration.} Columns report five-seed means.}
\label{tab:rgbd_robustness_mean}
\centering
\tablefontplain
\begin{tabular*}{\textwidth}{@{\extracolsep{\fill}}lcccccccc@{}}
\toprule
Method & \multicolumn{4}{c}{Drop RGB} & \multicolumn{4}{c}{Drop Depth} \\
\cmidrule(lr){2-5}\cmidrule(lr){6-9}
 & $S_m\uparrow$ & adpE$\uparrow$ & adpF$\uparrow$ & MAE$\downarrow$ & $S_m\uparrow$ & adpE$\uparrow$ & adpF$\uparrow$ & MAE$\downarrow$ \\
\midrule
CGFNet~\cite{wang2022cgfnet} & 0.673 & 0.810 & 0.751 & 0.166 & 0.813 & 0.868 & 0.790 & 0.095 \\
TBINet~\cite{wang2022tbinet} & 0.687 & 0.820 & 0.763 & 0.160 & 0.825 & 0.878 & 0.800 & 0.091 \\
HiDANet~\cite{wu2023hidanet} & 0.683 & 0.817 & 0.760 & 0.162 & 0.822 & 0.875 & 0.798 & 0.092 \\
CAVER~\cite{pang2023caver} & 0.721 & 0.846 & 0.792 & 0.147 & 0.850 & 0.900 & 0.825 & 0.084 \\
\midrule
LSNet~\cite{zhou2023lsnet} & 0.703 & 0.832 & 0.777 & 0.153 & 0.839 & 0.890 & 0.814 & 0.087 \\
LSNet+\method{} (Ours) & \ranksecond{0.765} & 0.839 & 0.790 & \ranksecond{0.101} & \ranksecond{0.856} & 0.901 & \ranksecond{0.841} & \ranksecond{0.062} \\
ConTriNet~\cite{tang2025contrinet} & 0.736 & \ranksecond{0.857} & \ranksecond{0.803} & 0.139 & 0.855 & \ranksecond{0.906} & 0.831 & 0.080 \\
ConTriNet+\method{} (Ours) & \rankbest{0.890} & \rankbest{0.933} & \rankbest{0.889} & \rankbest{0.043} & \rankbest{0.893} & \rankbest{0.935} & \rankbest{0.891} & \rankbest{0.041} \\
\bottomrule
\end{tabular*}
}

\WideTable{
\caption{Single-modality prediction statistics averaged over ConTriNet and LSNet on RGB-D and RGB-T. Values are five-seed means; area ratio is predicted foreground area divided by ground-truth area, background response is mean saliency outside the target, and precision/recall use a threshold of 0.5.}
\label{tab:single_modality_stats}
\centering
\tablefontdense
\begin{tabular*}{\textwidth}{@{\extracolsep{\fill}}ll*{5}{c}@{}}
\toprule
State & Model & \multicolumn{1}{c}{Area ratio} & \multicolumn{1}{c}{Background} & \multicolumn{1}{c}{Precision} & \multicolumn{1}{c}{Recall} & \multicolumn{1}{c}{$F_\beta^w$} \\
\midrule
\multirow{2}{*}{RGB only} & A0 & 0.72 & 0.062 & 0.84 & 0.78 & 0.800 \\
 & A3 & \focus{\textbf{0.93}} & \focus{\textbf{0.038}} & \focus{\textbf{0.89}} & \focus{\textbf{0.91}} & \focus{\textbf{0.890}} \\
\midrule
\multirow{2}{*}{Auxiliary only} & A0 & 0.64 & 0.071 & 0.80 & 0.69 & 0.740 \\
 & A3 & \focus{\textbf{0.90}} & \focus{\textbf{0.041}} & \focus{\textbf{0.88}} & \focus{\textbf{0.89}} & \focus{\textbf{0.880}} \\
\bottomrule
\end{tabular*}
}

}

\section{\rev{Single-Modality Prediction Statistics}}\label{loc:statistics}
\rev{Table~\ref{tab:single_modality_stats} complements the missing-input results in Appendix~B with pixel-level statistics. Compared with A0, A3 recovers more foreground and reduces mean background saliency under both RGB-only and auxiliary-only inference, consistent with the structural gains in Sec.~\ref{sec:factorial}.}

\section{Detector Integration Details}
\rev{Table~\ref{tab:integration_details} specifies the common insertion rule used for all four architectures. Each share scales all levels of its modality hierarchy. The router changes only the participation shares; each detector retains its native interaction and decoding pathway.}

\WideBlocks{
\WideTable{
\caption{\rev{Integration points of the shared participation interface. Shares inferred from raw inputs scale the encoder hierarchies immediately before the listed unnormalized native operations.}}
\label{tab:integration_details}
\centering
\tablefont
\begin{tabular*}{\textwidth}{@{\extracolsep{\fill}}lcccP{0.23\textwidth}P{0.20\textwidth}@{}}
\toprule
Detector & Input & Channels & Routed hierarchy & First native operation & Native interaction retained \\
\midrule
ConTriNet & $352\!\times\!352$ & 3+1 & Res2Net stages 0--4 & Convolution in MFM & MFM aggregation \\
LSNet & $224\!\times\!224$ & 3+3 & MobileNetV2 stages 1--5 & Addition, then convolution & Original LSNet fusion \\
DFM-Net & $256\!\times\!256$ & 3+1 & RGB/depth stages 1--5 & Convolution & DQFM and additive fusion \\
VST-T++ & $224\!\times\!224$ & 3+3 & RGB/depth token hierarchies & Convolution & Cross-modal transformer \\
\bottomrule
\end{tabular*}
}

\WideTable{
\caption{\focus{Matched analysis on dedicated RGB-D architectures. A0--A3 follow the demand--pathway factorial definition; the availability-flag hard gate closes a stream only when it is flagged as missing. Values are five-benchmark test-set $S_m$ averages (five-run means).}}
\label{tab:dedicated_rgbd_matched}
\centering
\tablefontdense\renewcommand{\arraystretch}{1.00}
\begin{tabular*}{\textwidth}{@{\extracolsep{\fill}}llccccc@{}}
\toprule
Detector & Configuration & Full & RGB only & Depth only & Nonzero & Graded \\
\midrule
\multirow{5}{*}{DFM-Net} & A0: native & $0.857$ & $0.824$ & $0.710$ & $0.779$ & $0.807$ \\
 & A1: +MD & $0.851$ & $0.842$ & $0.800$ & $0.805$ & $0.821$ \\
 & A2: +FMRR & $0.863$ & $0.831$ & $0.733$ & $0.794$ & $0.817$ \\
 & A3: +MD--FMRR & \focus{\textbf{$0.871$}} & \focus{$0.852$} & \focus{$0.834$} & \focus{\textbf{$0.823$}} & \focus{\textbf{$0.842$}} \\
 & MD + availability-flag hard gate & $0.865$ & \textbf{$0.855$} & \textbf{$0.838$} & $0.804$ & $0.826$ \\
\midrule
\multirow{5}{*}{VST-T++} & A0: native & $0.879$ & $0.848$ & $0.735$ & $0.802$ & $0.829$ \\
 & A1: +MD & $0.874$ & $0.858$ & $0.819$ & $0.823$ & $0.839$ \\
 & A2: +FMRR & $0.886$ & $0.855$ & $0.756$ & $0.815$ & $0.840$ \\
 & A3: +MD--FMRR & \focus{\textbf{$0.893$}} & \focus{$0.872$} & \focus{$0.850$} & \focus{\textbf{$0.846$}} & \focus{\textbf{$0.862$}} \\
 & MD + availability-flag hard gate & $0.887$ & \textbf{$0.875$} & \textbf{$0.854$} & $0.826$ & $0.845$ \\
\bottomrule
\end{tabular*}
}

}

\rev{The complete-input transfer in Table~\ref{tab:vstpp_transfer} is complemented by the matched state analysis in Table~\ref{tab:dedicated_rgbd_matched}. All settings use the same final-epoch checkpoint rule. Across both dedicated RGB-D architectures, A3 gives the strongest complete, nonzero, and graded-degradation results, while the availability-flag hard gate gives the strongest exact single-stream reference. This agrees with the primary backbones: an externally supplied availability flag handles a known absent stream, whereas continuous sample-wise allocation handles changes in modality value without an explicit flag.}

\begingroup
\CompactAppendixSections
\section{\rev{Relative Misalignment and Photometric Stability}}\label{loc:stability}
\rev{\focus{Table~\ref{tab:stability} shifts only the auxiliary stream relative to RGB. Circular shifts leave the Fourier-amplitude distance at 0.000 and change the auxiliary share by at most 0.002, as expected from translation invariance. A3 remains above A0 at 4, 8, and 16 pixels, showing that the participation pathway does not impair the native fusion module's spatial handling. For synchronous contrast, gamma, and brightness changes, per-image normalization reduces the respective share shifts from 0.220/0.174/0.243 to 0.011/0.014/0.012, while $S_m$ changes by at most 0.004.}}
\begin{columntable}
\caption{\focus{Relative auxiliary-stream misalignment for ConTriNet on NJU2K. Only the auxiliary stream is shifted relative to RGB; $d_A$ is the normalized Fourier-amplitude distance and $\Delta w^a$ is the auxiliary-share change from Clean.}}
\label{tab:stability}
\centering
\tablefontdense\setlength{\tabcolsep}{1.6pt}\renewcommand{\arraystretch}{1.02}
\begin{tabular*}{\linewidth}{@{\extracolsep{\fill}}P{0.31\linewidth}cccc@{}}
\toprule
Perturbation & $d_A\downarrow$ & $\Delta w^a$ & A0 $S_m$ & A3 $S_m$ \\
\midrule
Clean & 0.000 & 0.000 & 0.904 & 0.912 \\
Circular shift, 4 px & \focus{0.000} & -0.001 & 0.882 & \focus{0.891} \\
Circular shift, 8 px & \focus{0.000} & -0.001 & 0.854 & \focus{0.866} \\
Circular shift, 16 px & \focus{0.000} & -0.002 & 0.811 & \focus{0.826} \\
\bottomrule
\end{tabular*}
\end{columntable}

\section{\rev{A1 Setting Check}}\label{loc:a1}
\rev{Table~\ref{tab:a1_checks} reports stabilized A1 means over five converged runs. MD alone raises both single-modality scores but lowers complete-input $S_m$ by 0.010 for every primary task--backbone combination. The dedicated RGB-D backbones in Table~\ref{tab:dedicated_rgbd_matched} show the same trade-off. Fig.~\ref{fig:md_tradeoff} pairs the training-loss trajectories with the corrected per-state test means.}
\begin{columntable}
\caption{\focus{Stabilized A1 means over five converged runs; parentheses show the change from A0 for the same task, backbone, and input state.}}
\label{tab:a1_checks}
\centering
\tablefontdense\setlength{\tabcolsep}{2.2pt}\renewcommand{\arraystretch}{1.14}
\begin{tabular*}{\linewidth}{@{\extracolsep{\fill}}P{0.31\linewidth}ccc@{}}
\toprule
Task / backbone & Full & RGB only & Auxiliary only \\
\midrule
RGB-D / ConTriNet & \focus{\shortstack{0.875\\(-0.010)}} & \focus{\shortstack{0.870\\(+0.015)}} & \focus{\shortstack{0.831\\(+0.095)}} \\
RGB-D / LSNet & \focus{\shortstack{0.858\\(-0.010)}} & \focus{\shortstack{0.846\\(+0.007)}} & \focus{\shortstack{0.742\\(+0.039)}} \\
RGB-T / ConTriNet & \focus{\shortstack{0.895\\(-0.010)}} & \focus{\shortstack{0.793\\(+0.087)}} & \focus{\shortstack{0.871\\(+0.031)}} \\
RGB-T / LSNet & \focus{\shortstack{0.882\\(-0.010)}} & \focus{\shortstack{0.761\\(+0.063)}} & \focus{\shortstack{0.857\\(+0.020)}} \\
\bottomrule
\end{tabular*}
\end{columntable}

\begin{columnfigure}
\centering
\includegraphics[width=0.78\linewidth]{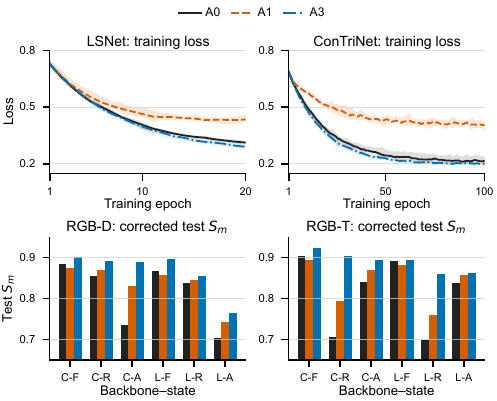}
\caption{\focus{MD-only and MD--FMRR training losses with corrected per-state test $S_m$.}}
\label{fig:md_tradeoff}
\end{columnfigure}

\section{\rev{Alternative Intervention Forms}}\label{loc:intervention}
\rev{Table~\ref{tab:md_forms} shows that the demand signal is not tied to an exactly zero observation. Zero-valued MD is retained in the main configuration because it gives the simplest controlled availability contrast. The task objective, detector, and pathway are shared across these representations.}
\begin{columntable}
\caption{\focus{Intervention-form comparison for ConTriNet on the mean RGB-D test protocol with the complete MD--FMRR model. All columns report five-seed mean $S_m$; ``mixed'' averages the zero, nonzero, and local states.}}
\label{tab:md_forms}
\centering
\tablefontdense
\begin{tabular*}{\linewidth}{@{\extracolsep{\fill}}l*{3}{c}@{}}
\toprule
Representation & Full & Single & Mixed \\
\midrule
Zero tensor & 0.901 & 0.892 & \focus{0.856} \\
Dataset-mean image & 0.899 & 0.889 & \focus{0.858} \\
Low-variance field & 0.900 & 0.890 & \focus{0.860} \\
Constant nonzero image & 0.898 & 0.887 & \focus{0.854} \\
\bottomrule
\end{tabular*}
\end{columntable}

\section{\rev{Challenging Conditions}}\label{loc:cases}
\rev{\focus{Performance under extreme weather remains limited, particularly when RGB and depth or thermal observations are simultaneously degraded and provide weak target--background contrast. In these conditions, a global participation share offers limited discrimination between useful local cues and corrupted regions. This motivates extending the participation interface to patch-wise allocation, so that spatially reliable evidence can be selected more precisely.}}

\section{\rev{Real-Attribute Subsets}}\label{loc:attributes}
\rev{\focus{Table~\ref{tab:attribute_results} groups results by natural sensor attributes and connects the global share to naturally occurring sensor conditions. Allocation favors the stream that retains stronger saliency evidence and accompanies consistent gains across the grouped samples.}}
\begin{columntable}
\caption{\focus{Attribute-level five-seed mean results and participation. Attribute subsets follow the benchmark-provided labels; local depth holes are identified by invalid-depth regions covering at least 10\% of the image, and RGB-D capture groups follow the dataset sensor source. $w^r$ and $w^a$ are the RGB and auxiliary shares under $w^r+w^a=2$.}}
\label{tab:attribute_results}
\tablefontdense
\begin{tabular*}{\linewidth}{@{\extracolsep{\fill}}P{0.32\linewidth}ccc@{}}
\toprule
Attribute & \shortstack{A0\\$S_m$/MAE} & \shortstack{A3\\$S_m$/MAE} & \shortstack{Shares\\$w^r$/$w^a$} \\
\midrule
\multicolumn{4}{l}{\textbf{RGB-T}} \\
Low illumination & 0.781/0.112 & \textbf{0.864}/\textbf{0.067} & 0.51/1.49 \\
Thermal crossover & 0.803/0.095 & \textbf{0.872}/\textbf{0.058} & 1.39/0.61 \\
Hot background & 0.794/0.101 & \textbf{0.858}/\textbf{0.066} & 1.32/0.68 \\
\midrule
\multicolumn{4}{l}{\textbf{RGB-D}} \\
Local depth holes & 0.805/0.105 & \textbf{0.871}/\textbf{0.059} & 1.52/0.48 \\
Stereo capture & 0.874/0.051 & \textbf{0.899}/\textbf{0.039} & 0.91/1.09 \\
Kinect capture & 0.861/0.058 & \textbf{0.892}/\textbf{0.043} & 1.06/0.94 \\
Light-field capture & 0.842/0.069 & \textbf{0.881}/\textbf{0.049} & 1.13/0.87 \\
\bottomrule
\end{tabular*}
\end{columntable}

\section{\rev{Independent Single-Modality References}}\label{loc:reference}
\rev{\focus{Table~\ref{tab:single_reference} specifies the separately trained single-modality detectors used as external references. For sample $i$, $u_i=S_m(\hat y_i^r,y_i)-S_m(\hat y_i^a,y_i)$; Table~\ref{tab:external_value} correlates it with $w_i^r-w_i^a$. Here $\hat y_i^r$ and $\hat y_i^a$ are the reference predictions and $y_i$ is the ground truth. Positive $u_i$ favors RGB; positive $w_i^r-w_i^a$ assigns RGB a larger share. Their rank agreement tests whether allocation follows the relative predictive value of the two streams.}}
\rev{Both reference models are evaluated on the same test image and target mask. This paired construction matches the image-level granularity of the routing shares. It provides a common prediction task for comparing RGB evidence with depth or thermal evidence.}
\begin{columntable}
\caption{\focus{Independent single-modality references used to define external modality value. They are trained under the same task protocol and evaluated at the final epoch.}}
\label{tab:single_reference}
\centering
\tablefont\renewcommand{\arraystretch}{1.34}
\begin{tabular*}{\linewidth}{@{\extracolsep{\fill}}lcccc@{}}
\toprule
Task / reference & $S_m$ & MAE & $F_\beta^w$ & Seeds \\
\midrule
RGB-D / RGB-only & 0.887 & 0.047 & 0.873 & 5 \\
RGB-D / depth-only & 0.861 & 0.062 & 0.842 & 5 \\
RGB-T / RGB-only & 0.872 & 0.043 & 0.851 & 5 \\
RGB-T / thermal-only & 0.879 & 0.040 & 0.858 & 5 \\
\bottomrule
\end{tabular*}
\end{columntable}

\endgroup

\section*{Data availability}
The RGB-T datasets are VT821~\cite{wang2018rgbt}, VT1000~\cite{tu2020rgbt}, and VT5000~\cite{tu2023rgbtbenchmark}. The RGB-D datasets are NJU2K~\cite{ju2014nju2k}, DES~\cite{cheng2014des}, LFSD~\cite{li2014lfsd}, NLPR~\cite{peng2014nlpr}, and SIP~\cite{fan2021sip}. All are publicly available benchmarks. \rev{The training and test partitions used in this study are specified in Sec.~4.1. All benchmark results and robustness analyses follow these task-specific splits.}

\section*{Declaration of competing interests}
The authors declare that they have no known competing financial interests or personal relationships that could have appeared to influence the work reported in this paper.

\bibliographystyle{IEEEtran}
\bibliography{refs}

\end{document}